\documentclass[lettersize,journal]{IEEEtran}
\usepackage{amsmath,amsfonts}
\usepackage{algorithmic}
\usepackage{algorithm}
\usepackage{array}
\usepackage[caption=false,font=normalsize,labelfont=sf,textfont=sf]{subfig}
\usepackage{textcomp}
\usepackage{stfloats}
\usepackage{url}
\usepackage{verbatim}
\usepackage{graphicx}
\usepackage{cite}
\usepackage{xcolor}
\usepackage{amsmath,amssymb}
\usepackage{multicol}

\usepackage{caption}
\usepackage[pagebackref=true,breaklinks=true,colorlinks,bookmarks=false]{hyperref}
\usepackage{booktabs}
\usepackage{array}
\newcolumntype{C}[1]{>{\centering\arraybackslash}m{#1}}

\newcommand{\etal}{\textit{et al}.~}
\newcommand{\ie}{\textit{i}.\textit{e}.}

\usepackage{dsfont}
\def\x{\times}
\newcommand{\blue}[1]{#1}
\newcommand{\tablestyle}[2]{\setlength{\tabcolsep}{#1}\renewcommand{\arraystretch}{#2}\centering\footnotesize}

\newlength\savewidth\newcommand\shline{\noalign{\global\savewidth\arrayrulewidth
  \global\arrayrulewidth 1pt}\hline\noalign{\global\arrayrulewidth\savewidth}}
  
\usepackage{tabulary,multirow,overpic}

\renewcommand{\star}{\textsuperscript{\textdagger}}
\newcommand{\doublestar}{\textsuperscript{\textdaggerdbl}}
\newcommand{\cmark}{\ding{51}}%
\newcommand{\xmark}{\ding{55}}%

\newcolumntype{P}[1]{>{\centering\arraybackslash}p{#1}}

\usepackage{pifont}
\usepackage{soul}
\renewcommand\st[1]{}

\begin{document}

\title{Pyramid Fusion Transformer for Semantic Segmentation}

\author{
Zipeng Qin, Jianbo Liu, Xiaolin Zhang, Aojun Zhou, Shuai Yi, Hongsheng Li

\thanks{Z. Qin, A. Zhou, and H. Li are with the Department of Electronic
Engineering, The Chinese University of Hong Kong.
E-mail: \{qinzipeng,aojunzhou\}@link.cuhk.edu.hk, hsli@ee.cuhk.edu.hk} 

\thanks{J. Liu, X. Zhang and Y. Shuai are with SenseTime Research.
E-mail: \{liujianbo,zhangxiaolin1,yishuai\}@sensetime.com}


}

\markboth{Journal of \LaTeX\ Class Files,~Vol.~14, No.~8, August~2021}%
{Shell \MakeLowercase{\textit{et al.}}: A Sample Article Using IEEEtran.cls for IEEE Journals}


\maketitle
\begin{abstract}
The recently proposed MaskFormer~\cite{maskformer} gives a refreshed perspective on the task of semantic segmentation: it shifts from the popular pixel-level classification paradigm to a mask-level classification method. In essence, it generates paired probabilities and masks corresponding to category segments and combines them during inference for the segmentation maps. In our study, we find that per-mask classification decoder on top of a single-scale feature is not effective enough to extract reliable probability or mask. To mine for rich semantic information across the feature pyramid, we propose a transformer-based Pyramid Fusion Transformer (PFT) for per-mask approach semantic segmentation with multi-scale features. The proposed transformer decoder performs cross-attention between the learnable queries and each spatial feature from the feature pyramid in parallel and uses cross-scale inter-query attention to exchange complimentary information. We achieve competitive performance on three widely used semantic segmentation datasets. In particular, on ADE20K validation set, our result with Swin-B backbone surpasses that of MaskFormer's with a much larger Swin-L backbone in both single-scale and multi-scale inference, achieving 54.1 mIoU and 55.7 mIoU respectively. Using a Swin-L backbone, we achieve single-scale 56.1 mIoU and multi-scale 57.4 mIoU, obtaining state-of-the-art performance on the dataset. Extensive experiments on three widely used semantic segmentation datasets verify the effectiveness of our proposed method.
\end{abstract}


\section{Introduction}
\label{sec:intro}

The goal of semantic segmentation is to assign each pixel of an image with a semantic class label. Over the past decade, encoder-decoder based methods are the mainstream models to address this task~\cite{fcn,efficientfcn,deeplabV3,9246719,8269325}. They usually use convolution-~\cite{fcn,efpn} or transformer-based~\cite{attn,9580642} networks to produce dense predictions from deep features generated by encoder networks. Efforts have been made to either design stronger backbone encoders~\cite{swin,vit,resnet,9580642} or decoders~\cite{detr,ddetr,unept,setr,segformer,deeplabV3plus} for the dense prediction task.

\begin{figure*}[t]
\centering
\begin{center}
    \includegraphics[width=14.0cm]{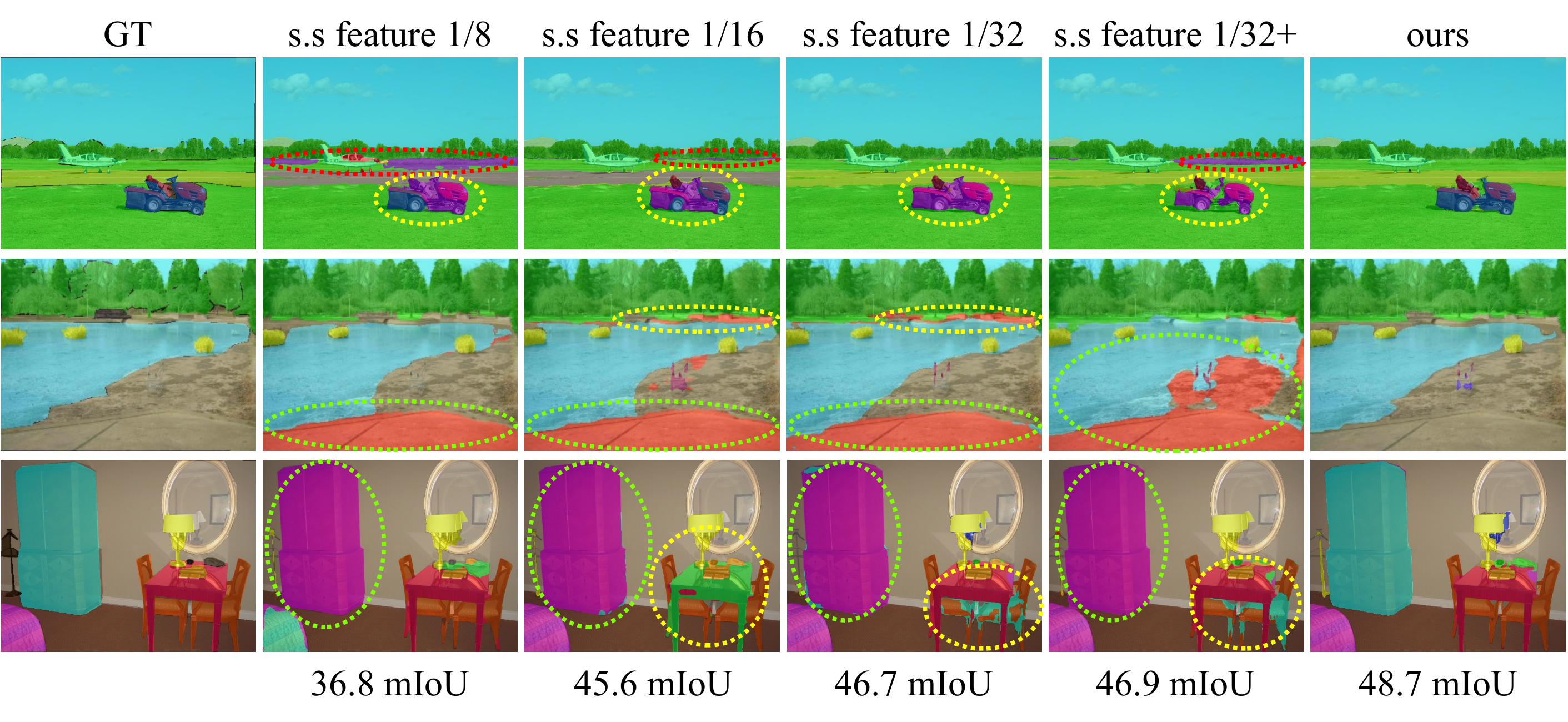} 
\end{center}
\caption{\textbf{Segmentation examples using different variants of the baseline}. We show the segmentation results from MaskFormer~\cite{maskformer} with a single-scale feature of 1/8, 1/16, and 1/32 the input resolution respectively, as well as a scaled version (s.s. feature $1/32+$) that matches the model capacity of our method. \blue{Along each row, the dashed circles show the deficiency in segmentation quality from MaskFormer using single-scale features, where the colors correspond to different regions in the images.}}
\label{fig:comparison_semseg}
\end{figure*}

Following FCN~\cite{fcn}, most of the encoder-decoder based semantic segmentation methods~\cite{deeplabV3,deeplabV3plus,efficientfcn,9546533} follow a per-pixel classification paradigm to address the task: the outputs of the models are spatial segmentation maps with categorical assignment for each pixel. The optimization happens at each pixel location, usually through a cross-entropy loss. Recently, Cheng~\etal took inspiration from DETR~\cite{detr} in object detection and proposed MaskFormer~\cite{maskformer} to advocate a path-breaking replacement for the per-pixel formulation. It adopts a novel per-mask classification method for semantic and panoptic segmentation. Instead of directly producing segmentation maps, a set of paired probabilities and masks are predicted corresponding to category segments in the input image. To this end, MaskFormer uses a transformer decoder to obtain probability-mask pairs with a set of learnable object queries using softmax activation and sigmoid activation respectively. Each object query is associated with a probability-mask pair and extracts useful information from encoder features through cross-attention~\cite{attn}. The probabilities and masks are end-to-end optimized through a classification loss and a mask loss separately. During inference, the probability-mask pairs are combined to produce the segmentation maps. 

\begin{table}[t]
  \centering
  
    \caption{MaskFormer~\cite{maskformer} with Swin-T backbone on top of single-scale and concatenated multi-scale feature maps for ADE20k validation set. $+$: we scale the original MaskFormer to match the model capacity of our method. 
    }
    
  \tablestyle{2pt}{1.2}\scriptsize\begin{tabular}{c|P{5em}|P{5em}|P{5em}|P{5em}|P{5em}|P{3em}}
   & s.s. feature 1/8 & s.s. feature 1/16 & s.s. feature 1/32~\cite{maskformer} & s.s. feature $1/32+$ & m.s. feature & Ours \\
  \shline
  FLOPS & - & - & 55G & 62G & 88G & \textbf{64G} \\
  \cline{1-7}
  params. & 42M & 42M & 42M & 67M & 67M & \textbf{63M} \\
  \cline{1-7}
  mIoU & 36.8 & 45.6 & 46.7 & 46.9 & 47.8 & \textbf{48.7} \\
  \end{tabular}

\label{tab:baseline}
\end{table}

Despite the MaskFormer's use of transformer module to integrate global information, we empirically observe that it still lacks the capability to produce fine details and is inclined to make contextual mistakes when decoding based on a single-scale feature map. We train MaskFormer with a Swin-T~\cite{swin} backbone on top of the spatial feature map of $1/8, 1/16$, $1/32$ input resolutions respectively (Tab.~\ref{tab:baseline} column 1, 2, and 3). As shown in Fig. \ref{fig:comparison_semseg}, it makes mistakes due to the lack of either fine-grained information or global context when decoding on a single-scale feature map.
However, neither increasing the model capacity nor using multi-scale features for the decoder is trivial or necessarily leads to an increase of segmentation accuracy (see Tab.~\ref{tab:baseline} column 4 and 5 and Sec.~\ref{sec:main}).
A naive attempt at using multi-scale features is to perform self-attention on the patch tokens of all scales. However, such an approach does not necessarily leads to performance gain and is computationally overwhelming as the computational cost increases quadratically with the number of tokens~(see Sec.~\ref{sec:ablations}).
We propose a simple yet elegant multi-scale transformer decoder, which efficiently propagates and aggregates multi-scale information from a feature pyramid for more accurate segmentation. Specifically, we use a set of learnable object queries and stacked transformer layers on top of \textit{each} multi-scale feature to produce probability-mask pairs parallelly and average the per-scale predictions for the final output.
The cross-scale communication is achieved via the proposed \textit{cross-scale inter-query attention} mechanism, which uses queries not only for information summarization within each scale but also for communication across scales. The small number of query tokens makes the cross-scale communication efficient.
Such a mechanism also allows queries of different scales to be aware of semantic information of other scales without directly computing on them.


\begin{figure*}[t]
\centering
\begin{center}
    \includegraphics[width=14.0cm]{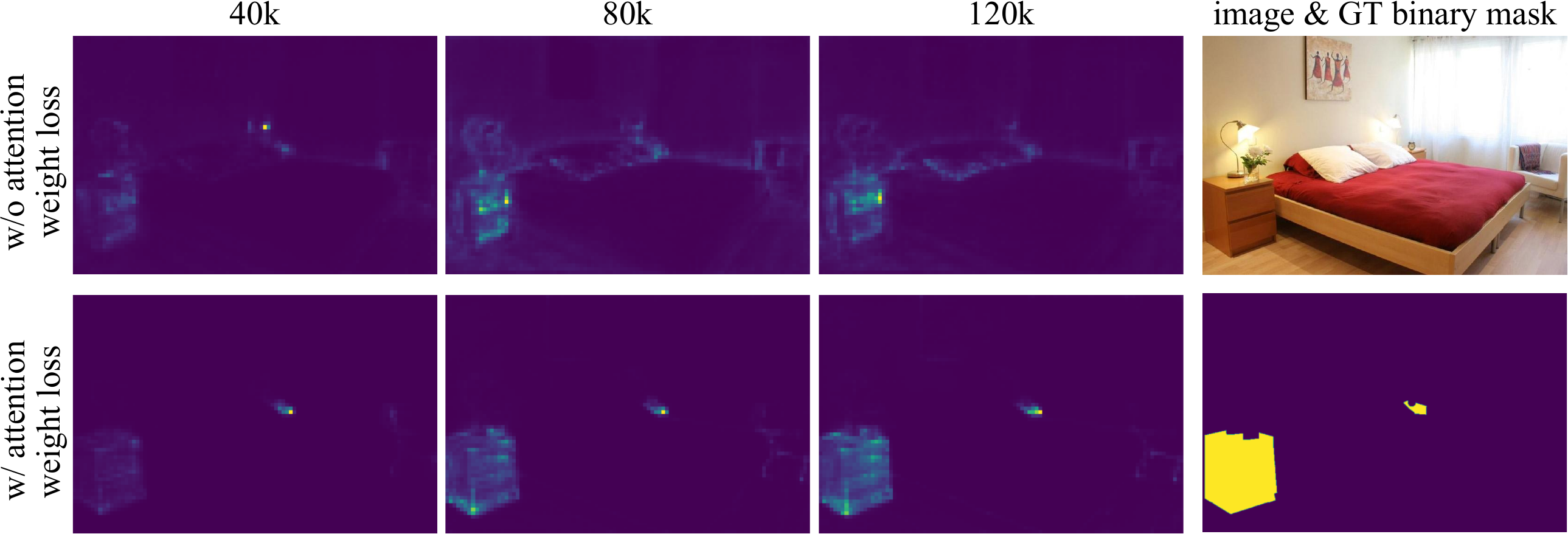} 
\end{center}
\caption{\textbf{Cross-attention weights and binary masks}. Visualizations obtained from the same model with and without our attention weight loss at different training iterations. Values are normalized to [0, 1] range. Images are from the ADE20K~\cite{ade20k} \textit{validation} set.}
\label{fig:attnw_vis}
\end{figure*}

In addition, we observe that the cross-attention weights are naturally correlated with the category segments. As shown in Fig.~\ref{fig:attnw_vis}, we visualize the cross-attention weights for the same object query at different training iterations and observe that they gradually concentration more on the concerned objects as the training progresses. Therefore, we aim at enhancing the correlation and propose a novel loss to directly supervise the cross-attention weights with the ground-truth category masks to facilitate and stabilize the training process. 

With the above proposed ingredients, we name our solution Pyramid Fusion Transformer (PFT), a segmentation transformer that efficaciously reasons from multi-scale feature maps with enhanced latent representations and consolidates predictions with high fidelity.
To demonstrate the effectiveness of our method, we conduct extensive experiments and ablation studies on three widely-used ADE20K~\cite{ade20k}, COCO-Stuff-10K~\cite{cocostuff}, and PASCAL Context~\cite{pascal} datasets and achieves state-of-the-art results on all the benchmarks. 

In summary, our contributions are threefold.
\begin{itemize}
    \item[\textbullet] We propose a Pyramid Fusion Transformer that effectively aggregates information from multi-scale feature pyramid to improve segmentation accuracy. The multi-scale communication and aggregation is efficiently achieved via a novel cross-scale inter-query attention mechanism.
    \item[\textbullet] Observing the strong correlation between the cross-attention map and the final segmentation map for each category, we apply a direct supervision on the cross-attention weights between class queries and spatial features.
    \item[\textbullet] Our proposed PFT achieves state-of-the-art performances on ADE20K and COCO-Stuff-10K datasets with extensive ablation studies validating the effectiveness of different ingredients.
\end{itemize}
\newcommand{\btext}[1]{\noindent\textbf{#1}}

\section{Related Works}
\label{sec:relatedworks}
\noindent {\bf Vision Transformer}. With the inspiration from success stories of transformer and attention mechanism~\cite{attn} in natural language processing tasks~\cite{bert}, the vision community enjoys a recent surge of interests in adapting the transformer structures into solving various vision tasks~\cite{vit,han2022dual,deit,detr,jiayao2022real}. The pioneering work ViT~\cite{vit} first proposes to divide an image input into a set of $16\times16$ patches, and sends them to a series of transformer layers consisting of self-attention modules and fully-connected layers with skip connections. On the task of image classification, it achieves competitive results with popular CNN networks such as ResNets~\cite{resnet}. Since the original ViT adopts an isotropic structure and can only produce feature maps of the same resolution, it is unsuitable for many downstream tasks that require multi-scale features. Moreover, the computational cost of patch tokens self-attention increases quadratically to the image size and quickly becomes overwhelming when ViT is applied to dense tasks that require high resolution inputs. To better adapt the transformer structure to dense vision tasks, researchers have designed various variants with stage-by-stage feature maps with shrinking resolutions~\cite{swin,9580642,p2t,cswin,fpt}. The improved transformers are both more efficient and accurate than the vanilla version when evaluated on downstream vision tasks, such as object detection and segmentation. Swin Transformer~\cite{swin} is arguably the most representative of these transformers with feature pyramid. In stead of performing self-attention on the entire set of patch tokens, it uses window partitions as constraints on where the attention mechanism is applied and allows communication among windows by using a shifted configuration. Interleaved patch merging module is used to reduce the feature resolution stage-by-stage. As a result, it achieves high performance in several downstream vision tasks with reduced computational costs and memory consumption.

\noindent {\bf Multi-scale feature representation}. As the semantic information at different scales can better capture different objects or structures, it is beneficial to utilize information across different feature scales~\cite{fpn,panopticfpn,efficientfcn,hattn}. \cite{fpn} shows that effectively processing multi-scale features generated from a hierarchical backbone network can improve performance for many segmentation and detection frameworks. Its success has inspired numerous works to adopt a pyramidal feature representation design and to mine helpful ingredients from multi-scale features~\cite{efficientfcn,zhou2021mffenet,ddetr,panoptic-segformer,fcos,efpn,panopticfpn,fapn}. For instance, \textit{EfficientFCN}~\cite{efficientfcn} encodes a set of semantic codewords by down-sampling all multi-scale features to 1/32 the input resolution to capture strong global context information from multi-scale features. It then produces assembly coefficients from concatenated features scaled to 1/8 resolution and linearly combine the codewords to produce the segmentation maps. Semantic FPN~\cite{panopticfpn} adds an extra top-down path that connect all levels in the feature pyramid to into a single 1/4 resolution feature map and generate segmentation through up-sampling. FaPN~\cite{fapn} further improves the fine-grained multi-scale representation by using deformable convolution~\cite{deconv} to better align feature fusion at each spatial location and achieves improved performance for segmentation tasks compared to FPN~\cite{fpn}.

\noindent {\bf Per-pixel classification semantic segmentation}. Starting from the seminal work Fully Convolutional Network~(FCN)~\cite{fcn}, encoder-decoder based per-pixel classification semantic segmentation framework has been the dominating paradigm for the task~\cite{deeplabV3,efficientfcn,gao2022fbsnet,psp,setr}. They usually encode an image into deep spatial features of different sizes using an encoder backbone network and use a decoder network to produce a spatial segmentation map of the same resolution as the input image. The encoder is usually a neural network pretrained on large-scale dataset~\cite{imagenet} and generates a feature pyramid rich with both fine-grained and contextual information. The decoder is another neural network that uses the multi-scale features to extract dense information and restore the original image resolution to predict a categorical label for each spatial location. During training, the encoder and decoder are end-to-end optimized, usually through a classification loss applied at each spatial location. Particularly, FCN~\cite{fcn} adopts as backbone encoder a CNN trained on ImageNet~\cite{imagenet} classification and discard the final fully connected layer to keep the spatial feature map. Deconvolution/up-sampling layers are used to generate segmentation maps with the same resolution as the input image. One of the recent works SegFormer~\cite{segformer} designs a transformer backbone that generates multi-scale features and uses an MLP decoder to predict segmentation results from the concatenated features. Despite the use of transformer architecture in its architecture, SegFormer still follows the per-pixel semantic segmentation paradigm that predicts and optimizes classification at each spatial location.

\noindent {\bf Per-mask classification semantic segmentation}. MaskFormer~\cite{maskformer} and Max-DeepLab~\cite{maxdeeplab} are among the pioneering works to use a mask classification paradigm in place of the end-to-end per-pixel classification segmentation approach. Contrary to FCNs that predict segmentation maps with per-pixel labelings, they produce paired results for masks and their corresponding class labels. To this end, they usually interact a set of object queries with the backbone features through stacked transformer layers and produce a probability-mask pair associated with each query. In particular, Max-DeepLab adds a transformer module to each backbone convolutional block and performs self/cross-attention to allow communications between the backbone and the decode head. MaskFormer extracts semantic information from 1/32 resolution feature through cross-attention between object queries and feature. The probabilities are generated by applying a linear layer on the queries and the masks are produced through dot-product between the queries and 1/4 resolution feature. Recently, Li~\etal~\cite{panoptic-segformer} proposes Panoptic Segformer, which shares a similar per-mask classification concept for panoptic segmentations on top of multi-scale feature maps. Specifically, it first applies a transformer decoder with sparse attention~\cite{ddetr} on the multi-scale features and uses the output queries to predict bounding box locations. The location-aware queries are then passed to a second transformer to perform dense cross-attention~\cite{attn} again with the multi-scale features. The predicted masks are then drawn from the attention weights between the queries and the spatial features and the probabilities are predicted in a separate branch. During inference, the masks and probabilities are combined to generate the segmentation maps.
\section{Method}
\label{sec:method}
\begin{figure*}[t]
  \centering
  \includegraphics[width=0.99\linewidth]{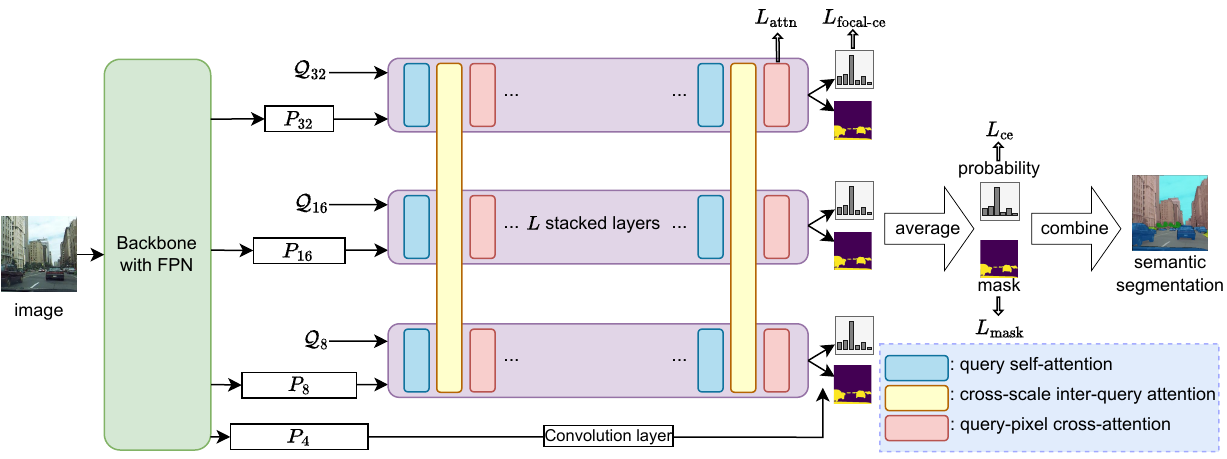}
  \caption{\btext{Overview of {Pyramid Fusion Transformer ({PFT})}} . {PFT} is composed of a backbone network with FPN and a set of parallel branches with \textit{cross-scale inter-query attention} to process the multi-scale features. We use $\mathcal{K}$ separate queries for each scale, same as the number of categories of each dataset.}
  \label{fig:arch0}
\end{figure*}

Our overall pipeline is shown in Fig.~\ref{fig:arch0}. Our proposed Pyramid Fusion Transformer (PFT) takes a feature pyramid encoded by a backbone network as input and adopts a novel multi-scale transformer decoder with a \textit{cross-scale inter-query attention} mechanism to efficiently fuse multi-scale information for accurate semantic segmentation. 
The backbone network, which can either be a convolutional or transformer network, receives an input image of size $H\times W \times 3$ and produces a hierarchy of feature maps $\{P_4, P_8, P_{16}, P_{32}\}$ by a Feature Pyramid Network (FPN) with a uniform channel dimension $C$ and spatial dimension $\frac{H}{s} \times \frac{W}{s}$, where $s=4,8,16,32$. Our PFT is applied on top of the three scales of feature maps $\{P_8, P_{16}, P_{32}\}$ as sequences of pixel tokens. Each scale has a separate set of $\mathcal{K}$ queries to estimate the confidences and locations of $\mathcal{K}$ semantic categories, where each query is only responsible for capturing semantic information of one assigned category. Within the transformer, we recurrently stack three types of attention layers: (1) an intra-scale query self-attention layer that conducts conventional self-attention between queries of the same scale, (2) a novel \textit{cross-scale inter-query attention} layer to efficiently communicate scale-aware information using the limited number of $3\mathcal{K}$ queries of the three different scales, and (3) an intra-scale query-pixel cross-attention layer that aggregates semantic information from flattened sequences of pixel tokens.



\label{sec:our_model}

\subsection{Pyramid Fusion Transformer with Cross-scale Inter-query Attention}
\label{sec:pft}


Multi-scale information is important for achieving accurate scene understanding.
Low-resolution feature maps are able to capture global context while high-resolution ones are better at discovering fine structures such as category boundaries~\cite{psp,panopticfpn,segfix,boundary}. It is therefore vital to propagate information across the multiple scales to capture both global and find-grained information. However, due to the high computational cost of directly applying attention on the large number of multi-scale pixel tokens, previous transformer-based semantic segmentation methods such as~\cite{unept,panoptic-segformer} often rely on the sparse attention mechanism on the pixel tokens to model pixel-to-pixel relations.
Contrary to their approaches, to avoid heavy computation, we propose to efficiently fuse the multi-scale information with our proposed \textit{cross-scale inter-query attention} mechanism. Instead of extracting global and local semantic information among the pixel tokens, we propose to fuse the multi-scale information in the query embedding space. Three types of attention layers are recurrently stacked in our PFT to achieve the goal.

\noindent \textbf{Intra-scale query self-attention.} Within each scale, the intra-scale query self-attention layer conducts self-attention only between category queries within the scale.
Specifically, for each scale, the $\mathcal{K}$ category queries $\mathcal{Q}_s \in \mathbb{R}^{\mathcal{K}\x C}$ with learnable positional encodings $\mathcal{P}_s \in \mathbb{R}^{\mathcal{K}\x C}$ are input into the layer.
\begin{align}
\label{eq:qselfattn}
\begin{split}
    &Q_s, K_s = \mathrm{Projection}(\mathcal{Q}_s + \mathcal{P}_s),~~~
    V_s = \mathrm{Projection}(\mathcal{Q}_s), \\
    &\mathcal{Q}_s =  \mathrm{Attention}(Q_s, K_s, V_s),~~~\mathrm{for}~s=8,16,32,
\end{split}
\end{align}
\noindent \blue{where $\mathrm{Projection}(\cdot)$ represents a linear projection layer and $\mathrm{Attention}(\cdot)$ is the multi-head attention module introduced in~\cite{attn}\footnote{\blue{We use the same notations to represent linear projection layers as $\mathrm{Projection}(\cdot)$ and multi-head attention module as $\mathrm{Attention}(\cdot)$ in the equations in the later parts of this paper as well.}}.} The category queries $\mathcal{Q}_s$ are learnable and randomly initialized before training. In the forward pass, they are updated by the stacked attention layers, while the learnable positional encodings $\mathcal{P}_s$ are shared at different depths.
Following~\cite{detr,ddetr,maskformer}, the positional encodings are only used to encode the embeddings $Q_s, K_s$ but not $V_s$ for self-attention. The self-attention is conducted only between queries from the same scale to obtain the updated queries ${\cal Q}_s$. Such an intra-scale self-attention layer consists of the commonly used multi-head self-attention sub-layer and Feed-Forward Network (FFN) sub-layer with layer normalization and residual connection.

\noindent \textbf{Cross-scale inter-query attention.}
As the above intra-scale self-attention is limited within each scale, their three sets of updated queries can only obtain information specific to each scale. To allow information propagation between the multiple scales for knowledge fusion, a novel \textit{cross-scale inter-query attention} layer is introduced. 
As the number of such scale-aware queries in each scale is much smaller than the number of all visual tokens in each scale, we achieve information propagation across the multiple scales by conducting  attention between the concatenated $3\mathcal{K}$ category queries $\mathcal{Q}_\text{all}=\mathrm{Concat}(\mathcal{Q}_8, \mathcal{Q}_{16}, \mathcal{Q}_{32}) \in \mathbb{R}^{3\mathcal{K}\x C}$ of the three scales.
The inter-query attention results are then sliced back to the queries of the three scales, which serve as the input for the follow-up intra-scale query-pixel cross-attention layer. To distinguish queries of different scales, we use the learnable positional encodings $\mathcal{P}_\text{all}=\mathrm{Concat}(\mathcal{P}_8,\mathcal{P}_{16},\mathcal{P}_{32})$ from the intra-scale query self-attention layer.
\begin{align}
\label{eq:xscaleattn}
\begin{split}
    &Q, K = \mathrm{Projection}(\mathcal{Q}_\text{all} + \mathcal{P}_\text{all}), ~~~
    V = \mathrm{Projection}(\mathcal{Q}_\text{all}), \\
    &\mathcal{Q}_8, \mathcal{Q}_{16}, \mathcal{Q}_{32} =  \mathrm{Attention}(Q, K, V) \\
\end{split}
\end{align}
\noindent where the attention outputs a sequence of category queries of length $3\mathcal{K}$ and are sliced to sub-sequences of length $\mathcal{K}$ and assigned to each $\mathcal{Q}_s$ for $s=8,16,32$. 
In this way, the cross-scale attention and information communication are efficiently achieved by using the small number of category queries.
The proposed cross-scale inter-query attention layer also consists of a multi-head attention sub-layer and an FFN sub-layer with layer normalization and residual connection following the classical design of the dot-product attention.

\noindent \textbf{Intra-scale query-pixel cross-attention.}
To aggregate dense pixel-level semantic information to the category queries, we conduct dense query-pixel cross-attention within each scale. Within each scale, the category queries $\mathcal{Q}_s$ perform cross-attention with the pixel tokens $P_s$ for $s=8,16,32$ using the multi-head attention.
\begin{align}
\label{eq:xattn}
\begin{split}
    &Q_s = \mathrm{Projection}(\mathcal{Q}_s + \mathcal{P}_s), \\
    &K_s = \mathrm{Projection}(P_s + \mathcal{P}_s^\text{sine}), \\
    &V_s = \mathrm{Projection}(P_s), \\
    &\mathcal{Q}_s =  \mathrm{Attention}(Q_s, K_s, V_s),~\mathrm{for}~s=8,16,32.
\end{split}
\end{align}
\noindent where the same learnable positional encodings $\mathcal{P}_s$ used in the above two attention layers are added to the category queries $\mathcal{Q}_s$, and fixed sinusoidal positional encodings $\mathcal{P}_s^\text{sine}$ are added to the pixel tokens $P_s$ following~\cite{detr}.

As illustrated by Fig.~\ref{fig:arch0}, in our pyramid fusion transformer, each transformer layer consists of the above three types of attention layers and is stacked for $L$ times to form an $L$-layer transformer. By default, we use $L=6$ layers.
For intra-scale query attention and intra-scale query-pixel cross-attention, we use separate weights for the linear projection layers, layer normalization, etc. for each scale.
The proposed {cross-scale inter-query attention} layers are placed between the intra-scale query self-attention layers and the intra-scale query-pixel cross-attention layers. The small number of category queries serve as the bridges to efficiently aggregate and propagate the pixel-level semantic information across the multiple scales.
Neither intra-scale nor cross-scale pixel-to-pixel attention is used in our PFT, avoiding heavy computational cost of dense multi-scale information fusion.


\vspace{2pt}
\noindent \textbf{Generating Segmentation Maps.} After the multiple transformer layers, the updated three sets of category queries $\{ \mathcal{Q}_s \in \mathbb{R}^{\mathcal{K} \x C} \}_{s=8,16,32}$ with multi-scale information can be used to generate the probability-mask pairs for segmentation.
To generate $\mathcal{K}$ probabilities of the $\mathcal{K}$ categories for the three scale, we use three separate linear projections to map the category queries $\{{\cal Q}_s\}_{s=8,16,32}$ at the output of our PFT to the three sets of $\cal K$ probability logits $\{\mathcal{L}_s^\text{prob} \in \mathbb{R}^{{\cal K}}\}_{s=8,16,32}$, each of the $\cal K$ logits representing the probability of the corresponding semantic category. \blue{The averaged logits $\mathcal{L}^\text{prob}$ of the three scales are followed by a sigmoid activation layer to generate the binary probabilities $p \in \mathbb{R}^{\cal K}$, which denotes the confidence of each category existing in the input image.}

\begin{align}
\begin{split}
&\mathcal{L}_s^\text{prob}=\mathrm{Linear}(\mathcal{Q}_s), \\
&\mathcal{L}^\text{prob}= \frac{\sum_{\text{all } s} \mathcal{L}_s^\text{prob}}{3}, \\
&p=\mathrm{\blue{sigmoid}}(\mathcal{L}^\text{prob}).
\end{split}
\label{eq:prob}
\end{align}

\blue{To generate the binary category masks for the multiple scales, we first apply a $3\x3$ convolutional layer on the backbone feature $P_4 \in \mathbb{R}^{C \x \frac{H}{4} \x \frac{W}{4}}$ to produce the mask feature $\mathcal{M} \in \mathbb{R}^{C \x \frac{H}{4} \x \frac{W}{4}}$. Each set of category queries $\mathcal{Q}_s$ for a scale $s$ then go through a multi-layer perceptron (MLP), where $\mathrm{MLP}(\mathcal{Q}_s)\in\mathbb{R}^{{\cal K} \times C}$. Then, they perform a matrix product with the mask feature $\mathcal{M}$ to produce the mask logits ${\cal L}_s^{\rm mask}\in \mathbb{R}^{{\cal K} \times \frac{H}{4} \times \frac{W}{4}}$.} The averaged logits $\mathcal{L}^\text{mask}$ of the three scales are followed by a sigmoid activation layer to obtain the mask probability maps $m \in \mathbb{R}^{{\cal K} \times \frac{H}{4} \times \frac{W}{4}}$ of each category.
\begin{align}
\begin{split}
    &\mathcal{L}_s^\text{mask} = \mathrm{MLP}(\mathcal{Q}_s) \otimes \mathcal{M}, \\
    &\mathcal{L}^\text{mask} = \frac{\sum_{\text{all }s} \mathcal{L}_s^\text{mask}}{3}, \\
    &m = \mathrm{sigmoid}(\mathcal{L}^\text{mask}),
\end{split}
\label{eq:mask}
\end{align}
where ${\cal L}_s^{\rm mask}$, ${\cal L}^{\rm mask}, \text{ and } m \in \mathbb{R}^{{\cal K} \times \frac{H}{4} \times \frac{W}{4}}$ and $\otimes$ denotes matrix multiplication. \blue{The matrix multiplication $\mathcal{L}_s^\text{mask}=\mathrm{MLP}(\mathcal{Q}_s) \otimes \mathcal{M}$ is performed between the $\mathcal{K}\times C$ tensor $\mathrm{MLP}(\mathcal{Q}_s)$ and the $C \times (\frac{H}{4} \times \frac{W}{4})$ tensor $\mathcal{M}$, and the result $\mathcal{L}_s^\text{mask}$ is therefore of shape ${\cal K} \times \frac{H}{4} \times \frac{W}{4}$. Note that we use sigmoid activation here since we do not enforce the masks of different categories to be exclusive to each other, following the practice in~\cite{maskformer}.}


For each category $k$'s probability and mask, $\{(p_k, m_k) |~p_k \in \mathbb{R},~m_k \in \mathbb{R}^{\frac{H}{4} \x \frac{W}{4}} \}_{k=1}^\mathcal{K}$, $p_k$ is the $k$-th entry (class) of $p$, $m_k$ is the $k$-th channel (class) of $m$.
The semantic segmentation map is obtained by probability-mask marginalization as that in MaskFormer~\cite{maskformer}, where the category prediction for pixel $(h,w)$ is computed as
\begin{align}
    {\rm ClassPrediction}(h,w) = \underset{k \in\{1,\dots,\mathcal{K}\}}{\rm argmax}~p_k \cdot m_k(h,w),
\label{eq:pred}
\end{align}
\noindent where the masks of different categories are weighted by the predicted probabilities of their existence. 

\subsection{Cross-attention Weight Loss for Stabilizing Training}
For an input image with $N$ categories present, we decompose the groundtruth segmentation map into $N$ groundtruth label-mask pairs. Since $N$ is usually smaller than the total number of semantic categories $\mathcal{K}$, we pad the ground-truth set with ``not-exist" category $\varnothing$. This results in a set of $\mathcal{K}$ padded groundtruth label-mask pairs $\{(c_k, m_k^\text{gt}) | ~c_k \in \{1,\dots,K,\varnothing\}, m_k^\text{gt} \in \{0,1\}^{H\times W} \}_{k=1}^{\mathcal{K}}$, where $c_k=k$ and the binary mask $m_k$ corresponds to all pixels belonging to the category if the $k$-th category is present in the image. If the $k$-th category is absent, $c_k=\varnothing$ and we do not have a groundtruth binary mask for it.

\noindent {\bf Cross-attention weight loss.} Recent studies~\cite{li2022exploiting,cdetr,smca} suggest that transformer-based models for vision dense tasks suffer from the difficulty of optimizing cross-attention between queries and spatial features and propose various ways to optimize cross-attention. We observe that the attention weights between the queries $\mathcal{Q}_s$ and spatial features $P_s$ for transformer-based per-mask semantic segmentation have strong correlation with the spatial arrangements of the semantic maps (see Fig.~\ref{fig:attnw_vis}). 
To better guide the query-pixel cross-attention during the training process, we explicitly enforce a novel attention weight loss on the cross-attention maps. Specifically, we obtain the attention weights $\mathcal{W}_s \in \mathbb{R}^{\mathcal{K} \x \frac{H}{s}  \x \frac{W}{s}}$ from the Attention operation in Eq.~\eqref{eq:xattn} 
\begin{align}
\label{eq:attnw}
    \mathcal{W}_s = {\rm softmax}(Q_s K_s^T /\sqrt{C}),~\mathrm{for}~s=8,16,32.
\end{align}
where $Q_s$ and $K_s$ are obtained from Eq.~\eqref{eq:xattn}. Each channel $w_k \in \mathbb{R}^{\frac{H}{s}  \x \frac{W}{s}}$ of $\mathcal{W}_s$ represents the spatial attention weight between the $k$-th query and the spatial pixel tokens. Here, we omit the multi-head attention notation for simplicity, for which we take the average of logits across all heads as the input to the softmax layer.

We normalize the groundtruth binary mask $m_k^{\text{gt}}$ to form a probability distribution~(\ie~values
sum to 1) as the supervision target. Additionally, we use a uniform probability~(\ie~all values equal $s^2/HW$) to supervise the non-existing category's attention weight. A cross-entropy loss naturally follows to fit to the target distribution.
\begin{align}
\label{eq:lossattn}
    L_{\text{attn}} = \lambda_\text{attn} L_\text{ce}(w_k, \mathrm{Norm}(m_k^{\text{gt}})),~\mathrm{for}~s=8,16,32
\end{align}
where we abuse the notation for groundtruth binary mask $m_k^{\text{gt}}$ a bit for absent categories. $\mathrm{Norm}(\cdot)$ is the normalization described above. $\lambda_\text{attn}$ is the loss weight. An extra $0.1$ weight multiplier is applied if $c_k=\varnothing$ to balance positive and negative samples. The final attention weight loss is averaged across all categories and scales.

We empirically find the formulation for non-existing categories' attention weights help reduce the variance of the performance. A more detailed experiment about our attention weight loss can be found in Sec.~\ref{sec:ablations}.

\vspace{2pt}
\noindent \textbf{Overall training loss.} Similar to MaskFormer, we use a mask loss $L_{\text{mask}}$ on the predicted masks $m$. The mask loss follows~\cite{maskformer} and consists of a binary focal loss~\cite{focal} $L_\text{focal}$ and a dice loss~\cite{dice} $L_\text{dice}$, where $L_{\text{mask}}=\lambda_\text{focal}L_\text{focal}+\lambda_\text{dice}L_\text{dice}$. $\lambda_\text{focal}$ and $\lambda_\text{dice}$ are hyperparameters balancing the two terms. Note that the mask loss is only applied at the averaged mask logits $\mathcal{L}^\text{mask}$ from the three scales and optimizes masks with groundtruth categories only. Masks corresponding to $\varnothing$ are simply discarded during training.

Our classification loss $L_{\text{cls}}$ consists of two terms: a binary cross-entropy loss $L_{\text{ce}}$ applied at the averaged probability logits $\mathcal{L}^\text{prob}$ from the three scales and a focal-style~\cite{focal} binary cross-entropy loss $L_{\text{focal-ce}}$ at each decoder outputs $\mathcal{L}_s^\text{prob}$ to adaptively reweight the hard samples following~\cite{focal}. The formulation of the focal-style cross-entropy is described in the supplementary. The classification loss $L_{\text{cls}}$ is a linear combination of the above two losses $L_{\text{cls}}=\lambda_\text{ce}L_\text{ce}+\lambda_\text{focal-ce}L_\text{focal-ce}$, where $\lambda_\text{ce}$ and $\lambda_\text{focal-ce}$ are hyperparameters balancing the two terms.

Our final training loss $L_\text{train}$ is a sum of the classification loss, the mask loss, and the attention weight loss,
\begin{equation}
\label{eq:trainingloss}
\begin{split}
 L_\text{train} = L_\text{cls} + L_\text{mask} + L_\text{attn}.
\end{split}
\end{equation}

Similar to DETR~\cite{detr} we apply supervision to each transformer layer's output queries. Besides, supervision is also applied to the input learnable queries $\mathcal{Q}_s$ before any transformer layer.
\section{Experiments}
\label{sec:experiment}

In this section, we demonstrate the effectiveness of our method with competitive semantic segmentation results and compare to both state-of-the-art per-pixel classification and mask-level classification frameworks on three popular segmentation datasets, ADE20K~\cite{ade20k}, COCO-Stuff-10K~\cite{cocostuff}, and PASCAL-Context~\cite{pascal}. We choose MaskFormer \cite{maskformer} as our baseline model because of its strong performance among the mask-level classification methods~\cite{maskformer,maxdeeplab,panoptic-segformer}.
In the ablations, we further study the effectiveness of our proposed components, including usage of multi-scale features, cross-scale inter-query attention, and the design of attention weight loss. Experimental results demonstrate that our model can learn useful information from multi-scale feature maps to deliver high quality segmentation maps with our proposed multi-scale transformer decoder and optimization for query-pixel cross-attention.

\subsection{Datasets and Implementation details}

\noindent \textbf{Datasets.} ADE20K~\cite{ade20k} is a semantic segmentation dataset with 150 fine-grained semantic categories, including \textit{thing} and \textit{stuff}. It contains 20,210 images for training, 2,000 images for validation and 3,352 images for testing. COCO-Stuff-10K~\cite{cocostuff} is a scene parsing dataset with 171 categories, not counting the class ``unlabeled". We follow the official split to partition the dataset into 9k images for training and 1k images for validation.
PASCAL-Context~\cite{pascal} contains pixel-level annotations for the whole scenes with 4,998 images for training and 5,105 images for testing. We evaluate our method on the commonly used 59 classes of the dataset.

\noindent \textbf{Implementation details.} We use the open-source segmentation codebase \textit{mmsegmentation}~\cite{mmseg} to implement PFT. We adopt Swin Transformer~\cite{swin} and ResNet~\cite{resnet} as backbone networks for evaluation. For ResNets, we report results obtained with ResNet-50 and ResNet-101, along with its slightly modified version ResNet-101c. ResNet-101c has its $7\x7$ stem convolution layer replaced by three consecutive $3\x3$ convolutions, which is a protocol widely adopted in semantic segmentation methods~\cite{deeplabV3plus,ma2018shufflenet,psp,deeplabV3,hu2017squeeze,he2019bag}. 

\noindent \textbf{Training settings.} Models are trained on ADE20K, COCO-Stuff-10K, and PASCAL-Context with 160k, 60k, and 40k-iteration schedules respectively. For the ADE20K dataset, $512\x512$ images are cropped after scale jittering, horizontal random flip, and color jittering. The same data augmentations are used for COCO-Stuff-10K dataset and PASCAL-Context dataset, while a crop size of $640 \x 640$ is used for COCO-Stuff-10K and $480 \x 480$ for PASCAL-Context. We use a batchsize of 16, 32, and 16 for ADE20K, COCO-Stuff-10K, and PASCAL Context respectively. Scale jittering is set to between 0.5 and 2.0 of the crop sizes. We set $\lambda_\text{attn}=0.1$ for our attention weight loss and detach the loss at 3/4 of the total training schedule (see Sec.~\ref{sec:ablations} for our analysis). We choose $\lambda_\text{focal-ce}$ from $\{1.0, 2.0\}$ for all datasets and provide the ablations of the loss weights. 
AdamW~\cite{adamw} is used as our optimizer with a linear learning rate decay schedule. For ResNet backbones, they are pretrained on ImageNet-1K and we use a learning rate of $10^{-4}$ and a weight decay of $10^{-4}$. The learning rate for Swin-Transformer backbones is set to $6\times 10^{-5}$ and a weight decay of $10^{-2}$ is used.
For Swin-T and Swin-S backbones, we use the official pretrained weights on ImageNet-1K~\cite{imagenet} with $224\x224$ resolution. For Swin-B and Swin-L, we use the official pretrained weights on ImageNet-22K with $384\x384$ resolution. All models are trained on a single compute node with 8 NVIDIA Tesla V100 GPUs. See the supplementary for a more details of hyperparameters and experimental settings.

\noindent \textbf{Evaluation settings.} We use mean Intersection-over-Union (mIoU) as our evaluation metric for semantic segmentation performance. Both the single-scale and multi-scale inferences are reported in our experiments. For multi-scale inference, we apply horizontal flip and scales of 0.5, 0.75, 1.0, 1.25, 1.5, and 1.75.

\subsection{Main results}
\label{sec:main}
\begin{table*}[t]
  \centering
  
  \caption{\btext{Experiments on ADE20K dataset}. Results reported on ADE20K \textit{validation} set. s.s.: single-scale inference. m.s.: multi-scale inference. \doublestar: backbones pretrained on ImageNet-22K. Improvements over the baseline model (MaskFormer) are reported in the \textcolor{gray}{gray} brackets.}
  
  \tablestyle{2pt}{1.3}\scriptsize\begin{tabular}{c|l|ccc|cccc}
  \shline
  & \multicolumn{1}{c|}{method} & backbone & crop size  & schedule & mIoU (s.s.) & mIoU (m.s.) & params. & FLOPS \\
  \hline
  \multirow{9}{*}{\rotatebox{90}{CNN}} 
  & OCRNet~\cite{ocrnet}
  & R101c                     & $520^2$ & 150k & - & 45.3 & - & - \\
  \cline{2-9}
  & GRAr~\cite{GRAr}
  & R101c                     & $544^2$ & 200k & - & 47.1 & - & - \\
  \cline{2-9}
  & \multirow{2}{*}{DeepLabV3+~\cite{deeplabV3plus}} 
  & \phantom{0}R50c           & $512^2$ & 160k & 44.0 & 44.9 & \phantom{0}44M & 177G \\
  & & R101c                   & $512^2$ & 160k & 45.5 & 46.4 & \phantom{0}63M & 255G \\
  \cline{2-9}
  & \multirow{3}{*}{MaskFormer~\cite{maskformer}} 
  & \phantom{0}R50\phantom{c} & $512^2$ & 160k & 44.5 & 46.7 & \phantom{0}41M & \phantom{0}53G \\
  & & R101\phantom{c}         & $512^2$ & 160k& 45.5 & 47.2 & \phantom{0}60M & \phantom{0}73G \\
  & & R101c                   & $512^2$ & 160k& 46.0 & 48.1 & \phantom{0}60M & \phantom{0}80G \\
  \cline{2-9}
  & \multirow{3}{*}{\textbf{PFT} (ours)} 
  & \phantom{0}R50\phantom{c} & $512^2$ & 160k& 45.5\textcolor{gray}{~(+1.0)} & 47.9\textcolor{gray}{~(+0.9)} & \phantom{0}62M & \phantom{0}61G \\
  & & R101\phantom{c}       & $512^2$ & 160k& 46.5\textcolor{gray}{~(+1.1)} & 48.4\textcolor{gray}{~(+1.2)} & \phantom{0}81M & \phantom{0}81G \\
  & & R101c                 & $512^2$ & 160k& \textbf{47.8}\textcolor{gray}{~(+1.8)} & \textbf{49.8}\textcolor{gray}{~(+1.7)} & \phantom{0}81M & \phantom{0}82G \\
  \hline\hline
  \multirow{10}{*}{\rotatebox{90}{Transformer}}
  & BEiT~\cite{beit}
  & ViT-L\doublestar    & $640^2$ & 160k & 56.7 & 57.0 & 441M & - \\
  \cline{2-9}
  & SETR~\cite{setr}
  & ViT-L\doublestar    & $512^2$ & 160k & 48.6 & 50.3 & 308M & - \\
  \cline{2-9}
  & \multirow{4}{*}{MaskFormer~\cite{maskformer}}
  &   Swin-T\phantom{\doublestar} & $512^2$ & 160k& 46.7 & 48.8 & \phantom{0}42M & \phantom{0}55G \\
  & & Swin-S\phantom{\doublestar} & $512^2$ & 160k& 49.8 & 51.0 & \phantom{0}63M & \phantom{0}79G \\
  & & Swin-B\doublestar & $640^2$ & 160k& 52.7 & 53.9 & 102M & 195G \\
  & & Swin-L\doublestar & $640^2$ & 160k& 54.1 & 55.6 & 212M & 375G \\
  \cline{2-9}
  & \multirow{4}{*}{\textbf{PFT} (ours)}
  &   Swin-T\phantom{\doublestar} & $512^2$ & 160k& 48.7\textcolor{gray}{~(+2.0)} & 50.1\textcolor{gray}{~(+1.3)} & \phantom{0}63M & \phantom{0}64G \\
  & & Swin-S\phantom{\doublestar} & $512^2$ & 160k& 51.0\textcolor{gray}{~(+1.2)} & 52.0\textcolor{gray}{~(+1.0)} & \phantom{0}84M & \phantom{0}87G \\
  & & Swin-B\doublestar & $640^2$ & 160k& 54.1\textcolor{gray}{~(+1.4)} & 55.7\textcolor{gray}{~(+1.8)} & 123M & 206G \\
  & & Swin-L\doublestar & $640^2$ & 160k& \textbf{56.1}\textcolor{gray}{~(+2.0)} & \textbf{57.4}\textcolor{gray}{~(+1.8)} & 232M & 385G \\
  \hline
  \end{tabular}

\label{tab:semseg:ade20k}
\end{table*}
\noindent \textbf{Results on ADE20K dataset.}
Tab.~\ref{tab:semseg:ade20k} summarizes our results on ADE20K {validation} set. We report both results from single-scale and multi-scale inferences. As shown in the table, when paired with the Swin-T backbone, PFT achieves 48.7 mIoU, improving over MaskFormer~\cite{maskformer} by 1.6 mIoU and matching the accuracy obtained by SETR~\cite{setr} with a much larger backbone. Notably, with the Swin-B backbone, we obtain an mIoU of 55.7, surpassing MaskFormer with a much larger Swin-L backbone. Our best model achieves a 57.4 mIoU, obtaining state-of-the-art performance without any bells-and-whistles~\cite{fapn,semask}. Consistent improvements with CNN backbones can also be observed from the results. 

\begin{table*}[t]
  \centering
  
  \caption{\btext{Experiments on COCO-Stuff-10K dataset}. Results reported on the \textit{validation} set. s.s.: single-scale inference. m.s.: multi-scale inference. \doublestar: backbones pretrained on ImageNet-22K. Results produced by our re-implementation are marked with \star. Improvements over MaskFormer are reported in the \textcolor{gray}{gray} brackets.}
  
  \tablestyle{2pt}{1.35}\scriptsize\begin{tabular}{c|l|ccc|ccc}
  \shline
   & \multicolumn{1}{c|}{method} & backbone & crop size & schedule & mIoU (s.s.) & mIoU (m.s.) & \#params. \\
  \hline
  \multirow{9}{*}{\rotatebox{90}{CNN}} 
  & OCRNet~\cite{ocrnet}
  & R101c                       & $520^2$  & \phantom{1}60k  &  -   & 39.5 & - \\
  \cline{2-8}
  & GRAr~\cite{GRAr}
  & R101c                     & $544^2$  & 100k & -    & 41.9 & - \\
  \cline{2-8}
  & \multirow{3}{*}{MaskFormer~\cite{maskformer}} 
  &   \phantom{0}R50\phantom{c} & $544^2$  & \phantom{1}60k & 37.1 & 38.9 & 44M \\
  & & R101\phantom{c}           & $640^2$ & \phantom{1}60k & 38.1 & 39.8 & 63M \\
  & & R101c                     & $640^2$ & \phantom{1}60k & 38.0 & 39.3 & 63M \\
  \cline{2-8}
  & \multirow{3}{*}{\textbf{PFT} (ours)} 
  & \phantom{0}R50\phantom{c}   & $640^2$  & \phantom{1}60k & 39.3\textcolor{gray}{~(+2.2)} & 40.6\textcolor{gray}{~(+1.7)} & 62M \\
  & & R101\phantom{c}           & $640^2$  & \phantom{1}60k & 39.6\textcolor{gray}{~(+1.5)} & 41.5\textcolor{gray}{~(+1.7)} & 81M \\
  & & R101c                     & $640^2$  & \phantom{1}60k & \textbf{40.9}\textcolor{gray}{~(+2.9)} & \textbf{43.0}\textcolor{gray}{~(+3.7)} & 81M \\
  \hline\hline
  \multirow{6}{*}{\rotatebox{90}{Transformer}}  
  & \multirow{3}{*}{MaskFormer~\cite{maskformer}}
  &   Swin-T\star\phantom{\doublestar} & $640^2$ & \phantom{1}60k & 42.2 & 42.5 & 42M \\
  & & Swin-S\star\phantom{\doublestar} & $640^2$ & \phantom{1}60k & 44.1 & 45.0 & 63M \\
  & & Swin-L\doublestar\star    & $640^2$ & \phantom{1}60k & 48.9 & 50.1 & 212M \\
  \cline{2-8}
  & \multirow{3}{*}{\textbf{PFT} (ours)}
  &   Swin-T \phantom{\doublestar} & $640^2$ & \phantom{1}60k & 43.0\textcolor{gray}{~(+0.8)} & 43.7\textcolor{gray}{~(+1.2)} & 63M   \\
  & & Swin-S \phantom{\doublestar} & $640^2$ & \phantom{1}60k & 44.8\textcolor{gray}{~(+0.7)} & 45.4\textcolor{gray}{~(+0.4)} & 84M  \\
  & & Swin-L\doublestar    & $640^2$ & \phantom{1}60k & \textbf{51.4}\textcolor{gray}{~(+2.5)} & \textbf{52.2}\textcolor{gray}{~(+2.1)} & 233M \\
  \hline
  \end{tabular}

\label{tab:semseg:coco}
\end{table*}

\begin{table*}[t]
  \centering
  \caption{\btext{Experiments on PASCAL-Context dataset}. Results reported on PASCAL-Context \textit{validation} set with 59 categories. s.s.: single-scale inference. m.s.: multi-scale inference. Results produced by our re-implementation are marked with $^{\text{\textdagger}}$. Improvements over MaskFormer are reported in the \textcolor{gray}{gray} brackets.}
  
  \tablestyle{2pt}{1.3}\scriptsize\begin{tabular}{c|c|ccc|ccc}
  \shline
  & \multicolumn{1}{c|}{method} & backbone & crop size & schedule & mIoU (s.s.) & mIoU (m.s.) & \#params. \\
  \hline
  \multirow{9}{*}{\rotatebox{90}{CNN}} 
  & \multirow{2}{*}{SFNet~\cite{sfnet}}
  & \phantom{0}R50c               & $512^2$ & \phantom{1}38k & - & 50.7   & - \\
  & & R101c                       & $512^2$ & \phantom{1}38k & - & 53.8   & - \\
  \cline{2-8}
  & \multirow{1}{*}{GRAr~\cite{GRAr}}
  &   R101c & $544^2$ & \phantom{1}50k & - & 55.7   & - \\
  \cline{2-8}
  & \multirow{3}{*}{MaskFormer~\cite{maskformer}} 
  &\phantom{0}R50\star\phantom{c} & $480^2$ & \phantom{1}40k & 52.5 & 54.1   & \phantom{0}44M \\
  & & R101\star\phantom{c}        & $480^2$ & \phantom{1}40k & 53.7 & 55.4   & \phantom{0}63M \\
  & & R101c\star                  & $480^2$ & \phantom{1}40k & 53.1 & 55.6   & \phantom{0}63M \\
  \cline{2-8}
  & \multirow{3}{*}{\textbf{PFT} (ours)} 
  & \phantom{0}R50\phantom{c}     & $480^2$ & \phantom{1}40k & 53.5\textcolor{gray}{~(+1.0)} & 55.0\textcolor{gray}{~(+0.9)} & \phantom{0}62M \\
  & & R101\phantom{c}             & $480^2$ & \phantom{1}40k & 54.7\textcolor{gray}{~(+1.0)} & 56.2\textcolor{gray}{~(+0.8)} & \phantom{0}81M \\
  & & R101c                       & $480^2$ & \phantom{1}40k & \textbf{55.2}\textcolor{gray}{~(+2.1)} & \textbf{57.3}\textcolor{gray}{~(+1.7)} & \phantom{0}81M \\
  \hline
  \end{tabular}

\label{tab:semseg:pascal}

\end{table*}
\noindent \textbf{Results on COCO-Stuff-10K dataset.}
We report our results on COCO-Stuff-10K dataset in Tab.~\ref{tab:semseg:coco}. As shown in the table, PFT obtains consistent improvements over MaskFormer with CNN backbones. A significant 3.7 mIoU improvement over MaskFormer is attained by using the ResNet-101c backbone under multi-scale inference. We train both the baseline and our model with the Swin-L backbone on COCO-Stuff-10K. As a result, we achieve 52.2 mIoU on the dataset, surpassing all previous state-of-the-art results on COCO-Stuff-10K and outperforming MaskFormer by 2.1 mIoU.

\noindent \textbf{Results on PASCAL-Context dataset.}
We present our results from PFT trained on PASCAL-Context dataset in Tab.~\ref{tab:semseg:pascal}. Our method beats MaskFormer with different backbone networks, showing steady improvements for the per-mask classification framework for semantic segmentation. Our most significant performance gain is obtained by the ResNet-101c backbone, which achieves 55.2 mIoU with only 18M additional parameters compared to MaskFormer, outperforming it by a 2.1 mIoU margin in single-scale testing. With multi-scale inference, we obtain 57.3 mIoU with ResNet-101c. To our knowledge, we have the best semantic segmentation performance with ResNet-101c backbone on PASCAL-Context dataset.

\begin{table*}[t]
  \centering
  
  \caption{\textbf{Comparison with Mask2Former~\cite{mask2former} on ADE20K dataset}. Results reported on ADE20K \textit{validation} set. s.s.: single-scale inference. m.s.: multi-scale inference. \doublestar: backbones pretrained on ImageNet-22K. Comparisons with Mask2Former~\cite{mask2former} are reported in the \textcolor{gray}{gray} brackets. MSAttn: stronger FPN based on multi-scale deformable attention from~\cite{mask2former}. FaPN~\cite{fapn}: a sophisticated FPN specially designed for semantic segmentation. All Mask2Former models use MSAttn FPN by default~\cite{mask2former}.}
  
  \tablestyle{1pt}{1.3}\scriptsize\begin{tabular}{l|ccc|cccc}
  \shline
  \multicolumn{1}{c|}{method} & backbone & crop size  & schedule & mIoU (s.s.) & mIoU (m.s.) & params. & FLOPS \\
  \hline
  \multirow{4}{*}{Mask2Former~\cite{mask2former}}
  &   Swin-T\phantom{\doublestar} & $512^2$ & 160k& 47.7 & 49.6 & \phantom{0}47M & \phantom{0}74G \\
  & Swin-B\doublestar & $640^2$ & 160k& 53.9 & 55.1 & 107M & 223G \\
  & Swin-L\doublestar & $640^2$ & 160k& 56.1 & 57.3 & 215M & 403G \\
  & Swin-L\doublestar + FaPN~\cite{fapn}& $640^2$ & 160k& 56.4 & 57.7 & 217M & - \\
  \cline{1-8}
  \multirow{4}{*}{\textbf{PFT} (ours)}
  &   Swin-T\phantom{\doublestar} & $512^2$ & 160k& 48.7\textcolor{gray}{~(+1.0)} & 50.1\textcolor{gray}{~(+0.5)} & \phantom{0}63M & \phantom{0}64G \\
  & Swin-B\doublestar & $640^2$ & 160k& 54.1\textcolor{gray}{~(+0.2)} & 55.7\textcolor{gray}{~(+0.6)} & 123M & 206G \\
  & Swin-L\doublestar & $640^2$ & 160k& 56.1\textcolor{gray}{~(+0.0)} & 57.4\textcolor{gray}{~(+0.1)} & 232M & 385G \\
  & Swin-L\doublestar + MSAttn & $640^2$ & 160k& \textbf{56.3}\textcolor{gray}{~(+0.2)} & \textbf{57.8}\textcolor{gray}{~(+0.5)} & 232M & 403G \\
  \hline
  \end{tabular}

\label{tab:supp-ade20k}
\end{table*}
\noindent \textbf{Compare to concurrent work.}
Mask2Former~\cite{mask2former} was recently proposed as a multi-scale variant for MaskFormer~\cite{maskformer}. Compared to our approach, Mask2Former extracts semantic information from multi-scale feature maps by using cross-attention in a round-robin fashion, \ie~the queries attend to the spatial features one by one in the cross-attention layers. Compared to our method, it uses a more sophisticated masked attention module based on predicted mask priors, and a stronger FPN variant based on multi-scale deformable attention~\cite{ddetr}, along with several other improvements~\cite{mask2former}. We compare the results obtained by~\cite{mask2former} to ours in Tab.~\ref{tab:supp-ade20k}. PFT achieves competitive performance with Mask2Former with fewer FLOPs even if Mask2Former uses a stronger FPN based on multi-scale deformable attention~\cite{ddetr}. Our FPN follows the conventional design with lateral connections and group convolutions. In particular, PFT with the Swin-T~\cite{swin} backbone outperforms Mask2Former by a 1.0 single-scale mIoU margin, while using $10$ GFLOPs fewer than Mask2Former. When we replace the FPN with the one used by Mask2Former, we obtain 56.3 single-scale mIoU and 57.8 multi-scale mIoU with the Swin-L backbone, surpassing Mask2Former's result and matches that of Mask2Former with FaPN~\cite{fapn}, a sophisticated Feature Pyramid Network specially designed for semantic segmentation.

\begin{table}[t]
  \centering
        \caption{\textbf{Experiments on ADE20K \textit{testing} set}. Results reported use multi-scale inference. MSAttn: stronger FPN based on multi-scale deformable attention from~\cite{mask2former}. FaPN~\cite{fapn}: a sophisticated FPN specially designed for semantic segmentation. Note that all Mask2Former models use MSAttn FPN by default.}
        \tablestyle{5pt}{1.3}\scriptsize\begin{tabular}{l|c|ccc}
        \shline
        \multicolumn{1}{c|}{method} & backbone & P.A. & mIoU & score \\
        \hline
        SETR~\cite{setr} & ViT-L & 78.35 & 45.03 & 61.69 \\
        UperNet~\cite{swin} & Swin-L & 78.42 & 47.07 & 62.75 \\
        \multirow{2}{*}{Mask2Former~\cite{mask2former}}
        & Swin-L                    & 79.36 & 49.67 & 64.51 \\
        & Swin-L + FaPN~\cite{fapn} & \textbf{79.80} & 49.72 & 64.76 \\
        \cline{1-5}
        \multirow{3}{*}{\textbf{PFT} (ours)}
        & Swin-L          & 79.53 & 50.14 & 64.84 \\
        & Swin-L + MSAttn & 79.41 & 49.26 & 64.34 \\
        & Swin-L + FaPN~\cite{fapn} & 79.37 & \textbf{50.63} & \textbf{65.00} \\
        \hline
        \end{tabular}
    \label{tab:supp-ade20k_test}
\end{table}
\noindent \textbf{Results on ADE20K test set.}
To demonstrate the superior performance of our proposed PFT, we additionally report the results on the \textit{test} set of ADE20K~\cite{ade20k} dataset. We adopt the same training settings for the experiments on the \textit{validation} set, including pretrained checkpoints, input resolution, hyperparameters, etc., except that we train our models on the union of the \textit{training} and \textit{validation} sets of ADE20K as a common practice. The results are from submitting the multi-scale inference results on the \textit{testing} set to the official evaluation server. As shown in Tab.~\ref{tab:supp-ade20k_test}, our PFT without any specialized FPNs achieves higher segmentation score than that obtained by Mask2Former with FaPN~\cite{mask2former}~(64.84 vs. 64.76) and surpasses all previous state-of-the-art methods. Replacing FPN with FaPN~\cite{fapn} in our framework pushes the mIoU to 50.63 and the segmentation score to 65.00. To our knowledge, it is the only method obtaining over 65.00 segmentation score on the ADE20K \textit{testing} set with the Swin-L backbone.

\subsection{Ablation Studies and Analysis}
\label{sec:ablations}
To evaluate the effectiveness of the components in our multi-scale transformer decoder for semantic segmentation, we conduct ablations on the multi-scale design and the cross-scale inter-query attention layer.
Furthermore, we provide the analysis of our attention weight loss and the effects of the removal of the loss at various training iterations. Unless otherwise specified, for all the experiments, we train our model with the Swin-T backbone on ADE20K dataset and detach our attention loss at 3/4 the training schedule (120k iterations) and report the single-scale mIoU on the \textit{validation} set.

\noindent \textbf{Comparison with stronger baseline models.}
To further study the improvements brought by our framework, we conduct experiments with different variants of the baseline model with the same Swin-T backbone and compare our performance. Specifically, we first train MaskFormer~\cite{maskformer} with different single-scale features of spatial shapes $1/8,1/16,1/32$ the input image resolution respectively, where using a $1/32$ scale feature map corresponds to the original MaskFormer. As shown in Tab.~\ref{tab:baseline} columns 1, 2, and 3, single-scale feature is insufficient to produce higher segmentation quality for the per-mask segmentation framework. We further augment the baseline by increasing the number of transformer layers from 6 to 8 and double the hidden dimension in transformer. The augmented baseline model matches the parameters and FLOPs of our PFT with the Swin-T backbone. As can be seen from Tab~\ref{tab:baseline} column 4, the performance slightly increases but still lags a considerable margin behind our multi-scale approach. However, using multi-scale features is a non-trivial task: directly inputting concatenated multi-scale features into the baseline model achieves a 47.8 mIoU~(Tab.~\ref{tab:baseline} column 5), which is lower than our method with the same backbone by 0.9 mIoU and consumes a larger computation~($88$ GFLOPs vs. $64$ GFLOPs).

\begin{figure}
    \centering
    \includegraphics[width=70mm]{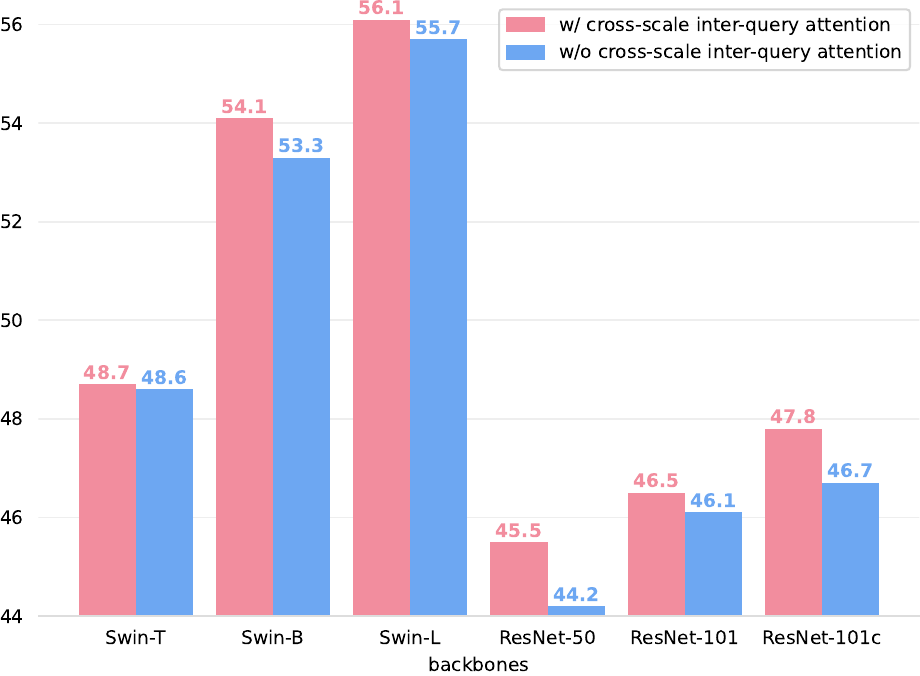}
    \captionof{figure}{\blue{\textbf{Ablations for cross-scale inter-query attention}. Results obtained on ADE20K \textit{validation} set.}}
    \label{fig:supp-qattn}
\end{figure}

\noindent \textbf{Effects of cross-scale inter-query attention.}
In our framework, we propose a novel cross-scale inter-query attention module to allow the multiple scales to propagate and aggregate useful information to other scales. To verify the benefit by such a module, we conduct experiments to remove the cross-scale query attention.
After such removal, our framework can be viewed as a multi-scale variant of MaskFormer~\cite{maskformer} with fixed-matching between queries and categories. 
\blue{As shown in Fig.~\ref{fig:supp-qattn}, with the Swin-T backbone, we observe a slight performance drop of 0.1 mIoU without the proposed module. With the Swin-B and the Swin-L backbones, the performance drops become more prominent~(0.8 and 0.4 mIoU drops respectively).
Moreover, significant performance drop of 1.3 mIoU is observed when paired with the ResNet-50 backbone. Noticeable performance drops are also observed with the ResNet-101 and the ResNet-101c backbones~(0.4 and 1.1 mIoU drops respectively). The comprehensive experimental results show that our proposed cross-scale inter-query attention module is beneficial for different backbones in general.}

\noindent \textbf{Efficiency of cross-scale inter-query attention compared to pixel tokens self-attention.}
\begin{table}[t]
  \centering
  \caption{\textbf{Efficiency of our cross-scale inter-query attention}. Measured on ADE20K. Training memory is the maximum memory of a single GPU. FLOPs of $512\times 512$ inputs. `IQA' denotes cross-scale inter-query attention. `Pixel SA' denotes pixel-token self-attention.}
  \tablestyle{6pt}{1.2}\scriptsize
  \begin{tabular}{c|c|c|c|c}
   Backbone & Attention Type & mIoU & Training Mem. & FLOPs \\
  \shline
  \multirow{3}{*}{Swin-T} & Ours w/o IQA & 48.6 & 13,762M & \phantom{0}62.4G \\
  & Ours w/ IQA & 48.7 & 13,795M & \phantom{0}63.7G \\
  & Ours w/ Pixel SA & 48.5 & 26,391M & 127.9G \\
  \shline
  \multirow{3}{*}{R50} & Ours w/o IQA & 44.2 & 18,233M & \phantom{0}60.1G \\
  & Ours w/ IQA & 45.5 & 18,264M & \phantom{0}61.5G \\
  & Ours w/ Pixel SA & 45.3 & 25,816M & 125.7G \\
  \hline
  \end{tabular}
\label{tab:pixel}
\end{table}
In our model, we propose to use the more efficient cross-scale inter-query attention module to propagate and aggregate information across scales instead of directly computing self-attention among the pixel tokens.
In Tab.~\ref{tab:pixel}, we add comparison between our proposed cross-scale inter-query attention module with 1) pixel-token self-attention, and 2) removing the inter-query attention. For 1), we replace our cross-scale inter-query attention with the pixel-token self-attention at every other transformer layer (inserting at all layers exceeds the GPU memory limit). The pixel-token self-attention between scales show no advantage over ours (Ours w/ IQA vs. Ours w/ Pixel SA) but requires more computational resources.
Compared with `Ours w/o IQA', `Ours w/ IQA' requires marginally additional resources but boosts performance.

\begin{figure*}[!t]
\begin{center}
    \includegraphics[width=13.0cm]{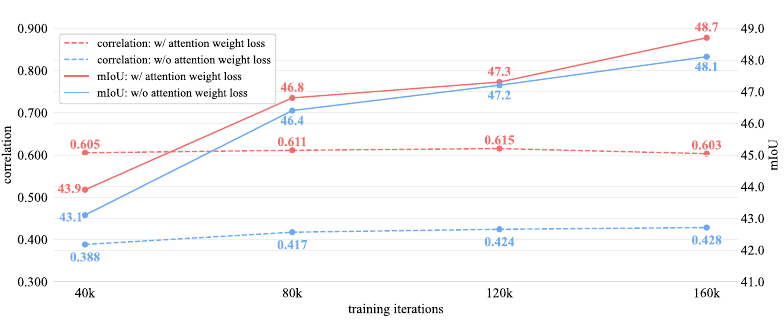} 
\end{center}
\caption{\textbf{Correlation between attention weights and category masks}. Left y-axis\&dashed lines: pearson correlation coefficient between attention weights and category masks. Right y-axis\&solid lines: mIoU at each training iteration.}
\label{fig:supp-corr_mIoU}
\end{figure*}

\begin{figure}
    \centering  
    \includegraphics[width=60mm,height=50mm]{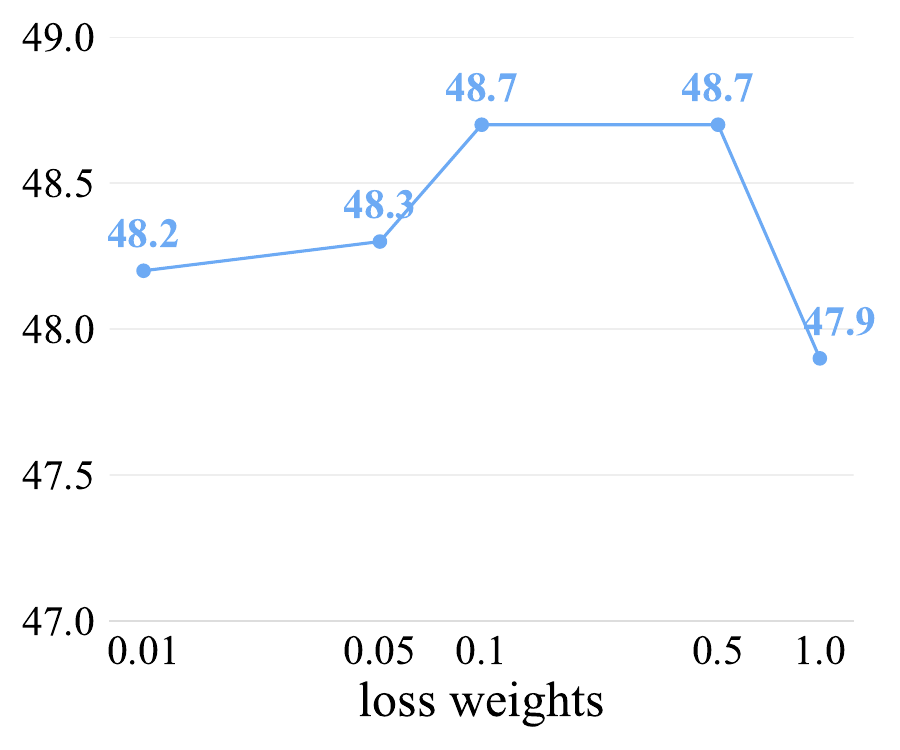}
    \captionof{figure}{\btext{Effects of different $\lambda_\text{attn}$} for attention weight loss. The x-axis is drawn on a log scale. We choose $\lambda_\text{attn}=0.1$ for all our experiments.}
    \label{fig:attnlossweight}
    \vspace{-6pt}
\end{figure}
\noindent \textbf{Attention weight loss.} We propose to use a novel attention weight loss to guide the cross-attention layer to focus on locations in the feature maps corresponding to the groundtruth category segments as well as optimizing the attention weights for non-existing categories with a uniform distribution. As shown in Fig.~\ref{fig:attnw_vis} row 2, using our attention weight loss helps the cross-attention to concentrate more on the regions corresponding to the categories. Besides, we calculate the pearson correlation coefficient between the attention weights and the binary masks to give a quantitatively analysis of the effects of our proposed loss. As shown in Fig.~\ref{fig:supp-corr_mIoU}~(dotted lines), our proposed loss significantly improves the correlation between attention weight maps and category masks at different training iterations. The improved correlations between attention weights and groundtruth binary masks lead to higher segmentation accuracy at different training iterations~(see Fig.~\ref{fig:supp-corr_mIoU} solid lines).

To further study the improvements from the proposed loss, we ablate the loss weight and verify our design through experiments. Fig.~\ref{fig:attnlossweight} shows the influences of $\lambda_\text{attn}$ and we choose $\lambda_\text{attn}=0.1$ for all our models. As shown in Tab.~\ref{tab:abla_attnloss} row 1, when the attention weight loss is not applied, the performance achieves 48.2 mIoU. If we do not optimize the non-existing categories' attention weights as part of the loss (Tab.~\ref{tab:abla_attnloss} row 2), the performance is improved yet with a larger standard deviation of 0.4. Finally, when the full version of our attention weight loss is applied, the performance is improved by 0.4 mIoU, obtaining 48.7 mIoU with a standard deviation of 0.1. We deduce that the additional supervision with uniform probability functions like a regularization technique and reduces the risk of non-existing categories' queries obtaining spurious semantic information from the features. 

We choose to remove the attention weight loss at certain training iteration so the model has better capacity of possibly learning information from regions outside the category segments. Tab.~\ref{tab:abla_attnloss} rows 3,  4, 5, and 6 show the effect of removing the loss at different stages of training. \blue{We notice that applying the loss throughout the entire training phase slightly degrades the performance and removing it at 120k iteration shows the best performance and variance~(Tab.~\ref{tab:abla_attnloss} rows 3\&6). The attention weight loss is used to modulate the optimization of the query-pixel cross-attention layers. Intuitively, if the loss is removed too early during training, such a modulation might show marginal effects on the final performance. On the other hand, applying this loss throughout the entire training phase might constrain too much the cross-attention that it fails to obtain beneficial semantic information outside the category regions.} We therefore choose 120k as the number of iterations optimizing the loss for ADE20K dataset and extend it to other datasets' training settings for our main experiments by matching the proportion of total training iterations (3/4 of the total iterations for each training schedule). \blue{In Fig.~\ref{fig:supp-corr_mIoU}, the slight decrease in correlation between the attention weights and category masks indicates that the queries indeed attend more to the regions outside the category masks after the loss removal. In the supplementary, we provide more visualizations and analysis of the proposed attention weight loss and the effect of the removal of the loss.}

\noindent \blue{\textbf{Supervision added to the learnable queries before any transformer layer.}
We conduct an ablation to remove this extra supervision. The performance drops slightly from 48.7 to 48.5 mIoU. We deduce that, even though the learnable queries haven't attained any information from backbone features before the transformer layers, adding this extra supervision can make them marginally work better with the network parameters, backbone features and prediction targets, since there are linear layers and spatial features involved when producing the predicted probability-mask pairs from the learnable queries~(see Eqs.~\ref{eq:prob} and~\ref{eq:mask}).}
\begin{table}[!t]
    \centering
    \caption{\btext{Analysis of query-pixel cross-attention weight loss}. $L_\text{attn}$ w/o: attention weight loss without optimizing attention weights from queries corresponding to non-existing categories in the input images. $L_\text{attn}$: our full attention weight loss. Each row is three runs of the same experiment.}
    \label{tab:abla_attnloss}
    \centering
    \tablestyle{2pt}{1.3}\scriptsize\begin{tabular}{c|c|cccc|c}
    \shline
    \multirow{2}{*}{$L_\text{attn}$ w/o} & \multirow{2}{*}{$L_\text{attn}$} & \multicolumn{4}{c|}{Stop Iteration} & \multirow{2}{*}{mIoU} \\
    \cline{3-6}
    & & 40k & 80k & 120k & 160k & \\
    \hline 
    \xmark & \xmark &        &        &        &        & $48.2 \pm 0.1$ \\
    \cmark &        &        &        & \cmark &        & $48.3 \pm 0.4$ \\
        & \cmark &        &        & \cmark &        & $48.7 \pm 0.1$ \\
        & \cmark & \cmark &        &        &        & $48.7 \pm 0.6$ \\
        & \cmark &        & \cmark &        &        & $48.4 \pm 0.7$ \\
        & \cmark &        &        &        & \cmark & $48.4 \pm 0.3$ \\
    \hline
    \end{tabular}
\end{table}

\noindent \blue{\textbf{The choice of multi-scale features.} We use multi-scale features of 1/8, 1/16, and 1/32 the input resolution as the inputs to our segmentation head. 
In Tab.~\ref{tab:multiscale}, we conduct experiments where we gradually add more multi-scale features to the models. Using three scales of features achieves the best trade-off between computation and performance. Reducing the number of multi-scale features noticeably degrades the performance, while adding an extra feature of 1/4 the input resolution brings too much computation overhead with marginal performance improvement.}
\begin{table}[t]
  \centering
  
    \caption{\blue{Ablation experiments for multi-scale features. We incrementally add features with larger resolutions to the models and report their performance on ADE20K \textit{validation} set. The FLOPs are calculated by using $512\times 512$ inputs. The first row shows what scales of multi-scale features are used in the segmentation head.}}
    
  \tablestyle{2pt}{1.2}\scriptsize\begin{tabular}{c|P{5em}|P{5em}|P{9em}|P{9em}}
   & 1/32 & 1/32+1/16 & 1/32+1/16+1/8~(ours) & 1/32+1/16+1/8+1/4 \\
  \shline
  FLOPS & 52G & 56G & \textbf{64G} & 88G \\
  \cline{1-5}
  mIoU & 47.0 & 48.1 & \textbf{48.7} & 48.9 \\
  \end{tabular}

\label{tab:multiscale}
\end{table}
\section{Conclusion}
\label{sec:conclusion}

We have presented Pyramid Fusion Transformer that aims to improve the segmentation performance in the per-mask classification semantic segmentation paradigm with multi-scale feature inputs. Using a transformer-based decoder with a novel cross-scale inter-query attention and optimization for query-pixel cross-attention, PFT shows steady improvements over MaskFormer on various datasets and achieves state-of-the-art results. We hope that our approach will inspire the community to further the research in improving per-mask classification segmentation framework.

\bibliographystyle{bibtex/IEEEtran}
\bibliography{egbib}

\end{document}


\title{Pyramid Fusion Transformer for Semantic Segmentation: Supplementary}

\markboth{Journal of \LaTeX\ Class Files,~Vol.~14, No.~8, August~2021}%
{Shell \MakeLowercase{\textit{et al.}}: A Sample Article Using IEEEtran.cls for IEEE Journals}


\maketitle

\begin{figure}[t!]
\section{Visualization and Analysis for Cross-Attention Weight Loss}
\centering
\begin{center}
    \includegraphics[width=8.5cm]{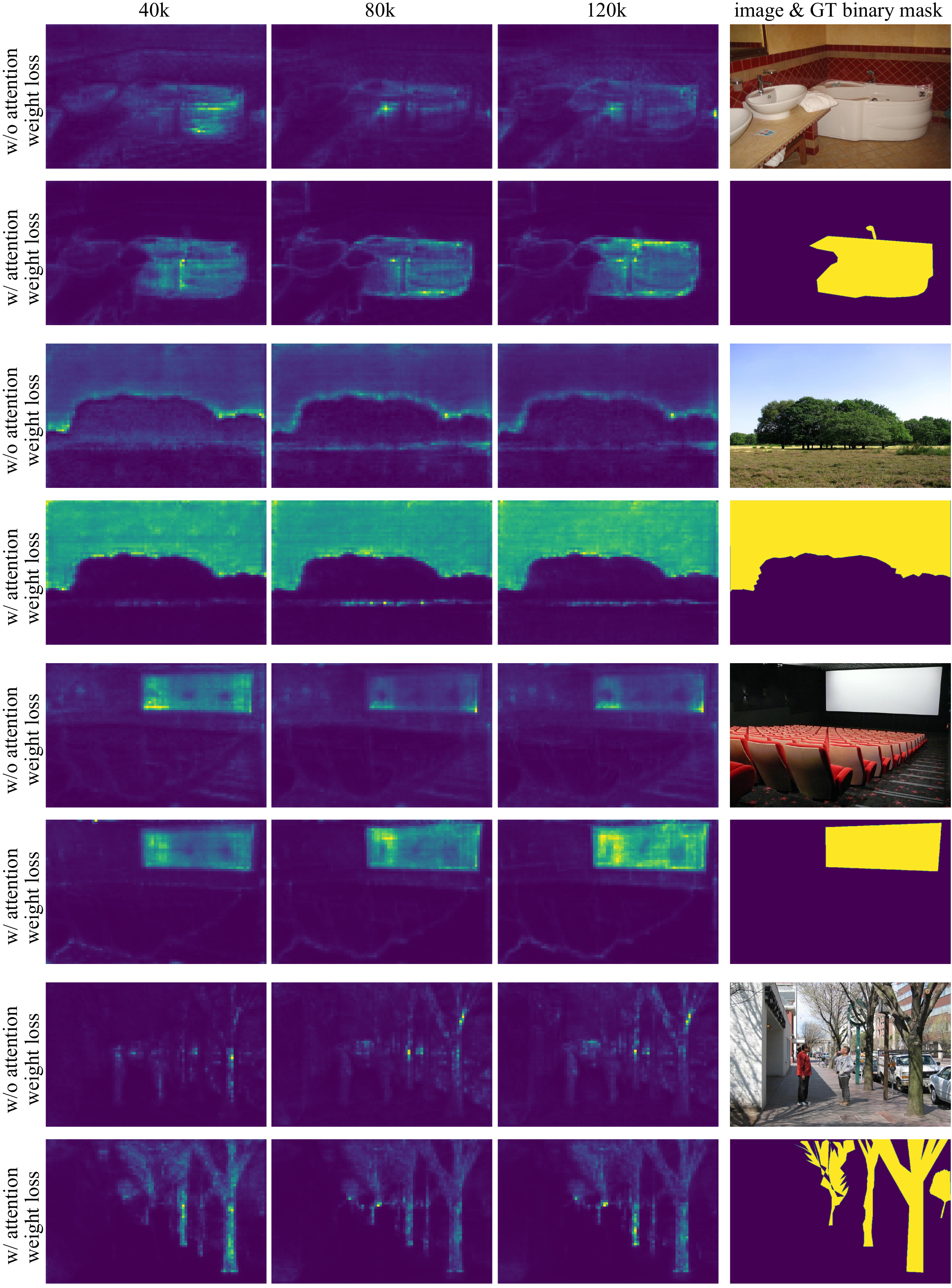} 
\end{center}
\caption{\textbf{Cross-attention weights and corresponding classes' binary masks}. Visualized attention maps are obtained from the same models with and without our attention weight loss at different training iterations. Values are normalized to $[0, 1]$. Images are from the ADE20K~\cite{ade20k} \textit{validation} set.}
\label{fig:supp-attnw_vis}
\vspace{-6pt}
\end{figure}

The query-pixel cross-attention is utilized to make the category queries aggregate useful semantic information from the spatial feature maps. We observe that the cross-attention maps are correlated to the ground-truth masks. To explicitly encourage the correlation, we propose a novel cross-attention weight loss $L_\text{attn}$ to apply additional supervisions to the cross-attention operation.  We visualize attention weights of our models trained with and without the proposed cross-attention loss\footnote{For this section, we report results based on checkpoints saved at 120k training iterations to better study the influences of the proposed loss.}. As shown in Fig.~\ref{fig:supp-attnw_vis}, we visualize the attention weight maps at 40k, 80k, and 120k training iterations respectively. Applying $L_\text{attn}$ makes the model more concentrate the cross-attention weights to the target regions.

\begin{figure}
  \centering
    \includegraphics[width=0.72\linewidth]{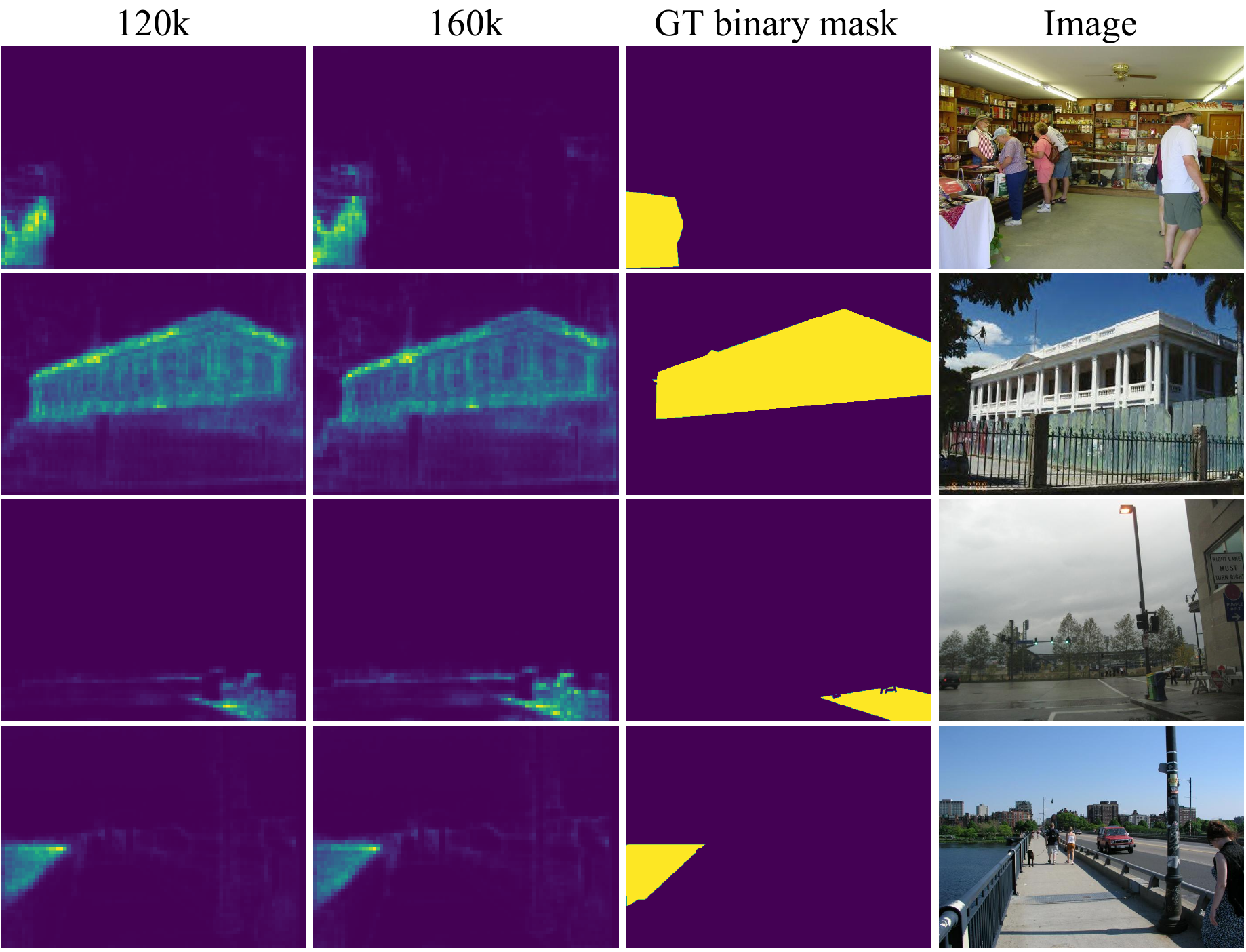}
  \caption{The attention weight loss is removed at 120k iteration. Visualized with Swin-T on ADE20K \emph{val}. Zoom in for better view.}
  \label{fig:removal}
\vspace{-6pt}
\end{figure}

We remove the attention weight loss at certain the training iteration to allow the model learns to obtain useful information outside the category segments. The slight drop in correlation coefficient indicates the queries indeed attend more to the regions outside the groundtruth masks after such removal~(Fig. 5, red dotted line, 120k vs. 160k). In addition, we included qualitative examples in Fig.~\ref{fig:removal} for illustration: the attention weights attend a bit more to the surrounding context of the ground truth category segments after removing the loss (120k vs. 160k iteration).
\section{Component Analysis}
\subsection{Logit Average vs. Prediction Average}
\label{supp:average}

In our framework, we use the averaged logits from three scales for both probability and mask predictions~(see Eqs.~(4-5) in the main paper). The final prediction for probabilities and masks are produced by a softmax layer and a sigmoid layer on the averaged logits, respectively. To validate our design choice, we train a model such that the final prediction is the averaged segmentation map from the three scales instead. To train this model, we apply the classification loss $L_\text{cls}$ and mask loss $L_\text{mask}$ on the three scales' logits before average and produce a segmentation map for each scale. The segmentation maps are produced similar to Eq.~(6) while the per-pixel argmax is applied on the averaged segmentation. The variant achieves 47.8 single-scale mIoU on the ADE20K~\cite{ade20k} \textit{validation} set, compared to 48.7 mIoU achieved by our logit averaging. 
This validates our design choice that the final segmentation predictions should be generated from the averaged logits from the three scales, not the averaged segmentations.

\subsection{Query Assignment with Fixed-matching vs. Bipartite-matching}

We propose to use each query to predict the probability-mask pair for each category. Therefore, a fixed-matching query-class assignment is adopted for the predictions and the categories. To compare this design choice with the bipartite-matching strategy used in~\cite{maskformer}, we train a variant of our model that uses bipartite-matching, where the probability logits in Eq.~(4) have $\mathcal{K}+1$ dimension corresponding to the $\mathcal{K}$ categories plus $\varnothing$ not-exist class. And the segmentation map generation pipeline follows~\cite{maskformer}. However, such an assignment between predictions and ground-truth masks makes it unclear how to average the logits produced from the three scales, since each query might predict for different categories across images~\cite{maskformer}. Therefore, for the bipartite-matching strategy, we adopt prediction averaging described in Sec.~\ref{supp:average}. As a result, we obtain 47.6 mIoU, which is similar to that of the variant we develop in Sec.~\ref{supp:average} for prediction space average. This is consistent with the observation made in~\cite{maskformer} that fixed-matching and bipartite-matching do not produce significant performance differences when models are trained under the same settings. Our choice of fixed-matching 
strategy enables optimization applied at the averaged logits and is more suitable for our semantic segmentation framework.
\section{Focal-Style Cross-Entropy Loss}
\label{sec:focalce}
\subsection{Formulation}

We use a focal-style~\cite{focal} cross-entropy loss $L_{\text{focal-ce}}$ as implemented in~\cite{multifocal} to supervise the probabilities predicted at each scale $s$. Specifically, within each scale $s$, we use a softmax layer on top of $\mathcal{L}_s^\text{prob}$ to generate probabilities $p_s \in \mathbb{R}^\mathcal{K}$ following Eq.~(4) in the main paper. For probability $p_{s,k}$ from the $k$-th position of $p_s$, we use $L_{\text{focal-ce}}(p_{s,k}) = - \alpha_t (1 - p_t)^\gamma \log p_t$, where

\vspace{-15pt}
\begin{align}
    \alpha_t &= 
    \begin{cases} 
        \alpha &\text{if $c_k \not= \varnothing$}\\
        1-\alpha &\text{otherwise.}
    \end{cases}
\end{align}
\vspace{-3pt}
\begin{align}
    p_t = 
    \begin{cases} 
        p_{s,k} &\text{if $c_k \not= \varnothing$}\\
        1 - p_{s,k} &\text{otherwise,}
    \end{cases}
\end{align}
\vspace{-5pt}

where $\alpha$ and $\gamma$ are hyperparameters used in focal loss~\cite{focal}. Following~\cite{focal}, we set $\alpha=0.25$ and $\gamma=2.0$ for all our experiments.

\subsection{Loss Weights Ablation}

\begin{figure}[!t]
    \centering  
    \includegraphics[width=60mm]{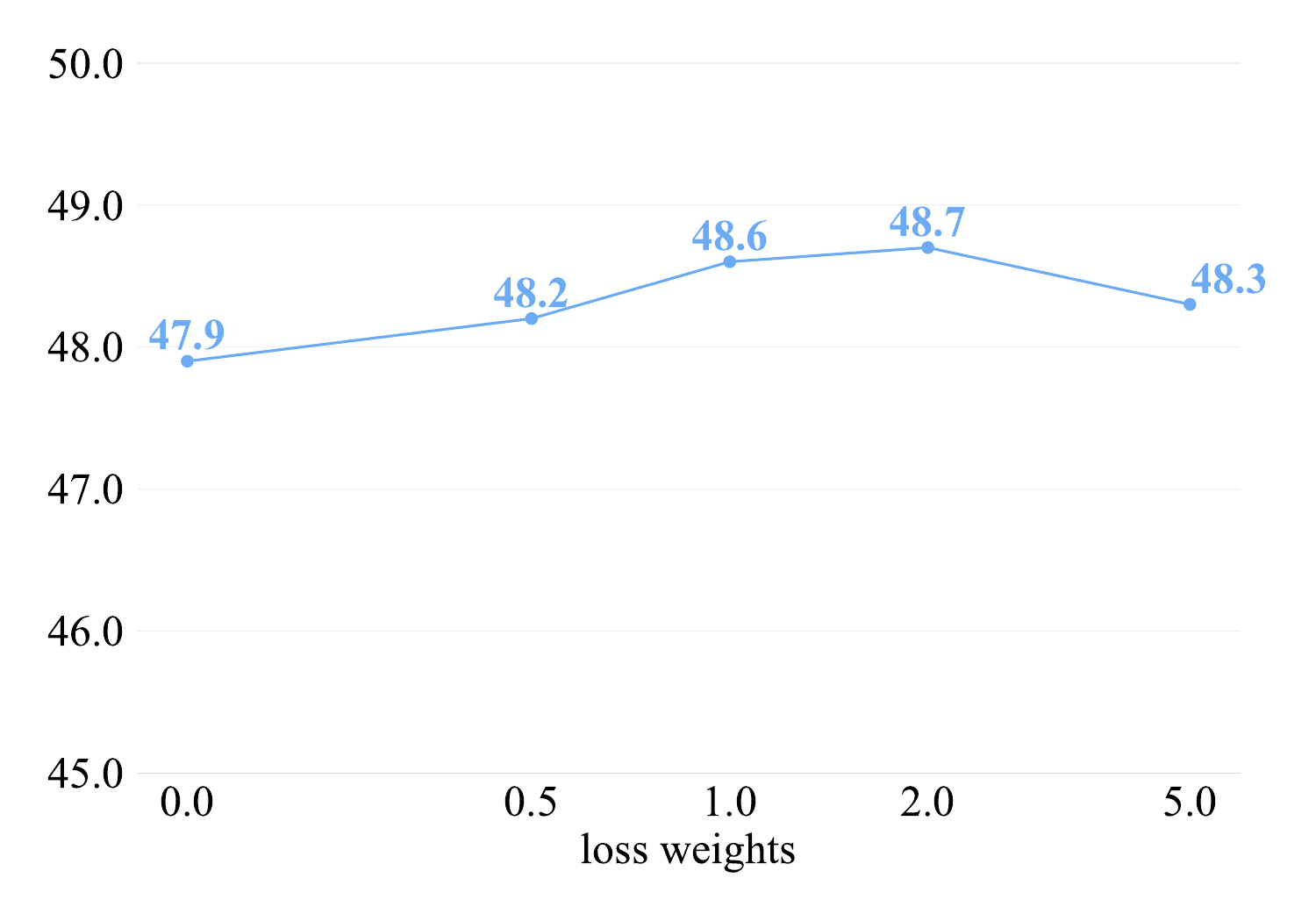}
    \captionof{figure}{\textbf{Loss weights for focal-style cross-entropy loss}. The x-axis is drawn on a log scale. We pick $\lambda_\text{focal-ce}\in \{1.0, 2.0\}$ as our candidate parameters given the results.}
    \label{fig:supp-focalw}
\end{figure}

For the focal-style cross-entropy loss, we use a $\lambda_\text{focal-ce}$ to balance the contribution of the loss to the training. Fig.~\ref{fig:supp-focalw} provides an quantitatively evaluation of the effects of different loss weights $\lambda_\text{focal-ce}$ on the segmentation accuracy. We use $\lambda_\text{focal-ce}\in \{0.0, 0.5, 1.0, 2.0,5.0\}$, where using a $0.0$ weight is equivalent to removing the loss. As shown in Fig.~\ref{fig:supp-focalw}, applying this additional loss term is 
beneficial. Using $\lambda_\text{focal-ce}=1.0 \text{~or~} 2.0$ gives a performance boost around 0.7 mIoU compared to not including the loss. Continuing increasing the loss weight does not lead to more growth in performance.

\begin{table}[t]
        \centering
	    \caption{\textbf{Hyperparameters and training settings for ADE20K~\cite{ade20k}, COCO-Stuff-10K~\cite{cocostuff}, and PASCAL-Context~\cite{pascal}}. We mostly follow the setups for hyperparameters reported in~\cite{maskformer}.}
	    \tablestyle{2pt}{1.3}\scriptsize\begin{tabular}{l|cc}
		\shline
		 & ~~ResNet & Swin \\
		\hline
		learning rate & ~~$1e^{-4}$ & $6e^{-5}$ \\
		\hline
		weight decay & ~~$1e^{-4}$ & $1e^{-2}$ \\
		\hline
		\# of decoder layers & \multicolumn{2}{c}{$6$} \\
		\hline
		\multirow{2}{*}{$L_\text{focal-ce}$} &  \multicolumn{2}{c}{$\lambda_\text{focal-ce}=1.0/2.0\text{ (Swin-T\&S)}$} \\
		& \multicolumn{2}{c}{$\alpha=0.25$, $\gamma=2.0$} \\
		\hline
		$\lambda_\text{ce}$~\cite{maskformer} & \multicolumn{2}{c}{$1.0/0.5\text{ (PASCAL Context)}$} \\
		\hline
		$\lambda_\text{dice}$~\cite{maskformer,dice} & \multicolumn{2}{c}{$1.0$} \\
		\hline
		$\lambda_\text{focal}$~\cite{maskformer,focal} & \multicolumn{2}{c}{$20.0$} \\
		\hline
		backbone lr multiplier & ~~$0.1$~~~ & $1.0/0.2\text{ (Swin-B\&L)}$ \\
		\hline
		\end{tabular}
	\label{tab:supp-hparams}
\end{table}

\section{Hyperparameters and Training Settings}

In this section we report our hyperparameters and experimental settings for the three datasets we use. We largely follow~\cite{maskformer} in designing our experiments and the default settings are recorded in Tab.~\ref{tab:supp-hparams}. We use the same dice loss~\cite{dice} and focal loss~\cite{focal} as in~\cite{maskformer}, with the same loss weights $\lambda_\text{dice}=1.0$ and $\lambda_\text{focal}=20.0$ respectively. We use $\lambda_\text{ce}=1.0$ for ADE20K and COCO-Stuff-10K and $\lambda_\text{ce}=0.5$ for PASCAL-Context. For focal-style cross-entropy loss's hyperparameters, we use the default $\alpha=0.25$ and $\gamma=2.0$~\cite{focal}. For Swin-T and Swin-S backbones, we use a $\lambda_\text{focal-ce}=2.0$ and we use $\lambda_\text{focal-ce}=1.0$ in other settings. Besides, we use a backbone learning rate multiplier of $0.1$ for CNN backbones and a multiplier of $0.2$ for the larger Swin-B and Swin-L backbones to avoid the risk of overfitting. From our experiments, we find that using a $\lambda_\text{ce}=0.5$ works best for PASCAL-Context~\cite{pascal} with both the baseline model and our method, possibly due to the smaller number of classes in the dataset.

\vspace{-3pt}
\bibliographystyle{bibtex/IEEEtran}
\bibliography{egbib}